\documentclass[letterpaper]{article} 
\usepackage{aaai23}  
\usepackage{times}  
\usepackage{helvet}  
\usepackage{courier}  
\usepackage[hyphens]{url}  
\usepackage{graphicx} 
\usepackage{natbib}  
\usepackage{caption} 
\usepackage{algorithm}
\usepackage{algorithmic}

\usepackage{newfloat}
\usepackage{listings}
\DeclareCaptionStyle{ruled}{labelfont=normalfont,labelsep=colon,strut=off} 
\floatstyle{ruled}
\newfloat{listing}{tb}{lst}{}
\floatname{listing}{Listing}
\usepackage{amsmath}
\usepackage{amsfonts}
\usepackage{caption}
\usepackage{subcaption}

\title{FLAME: Free-form Language-based Motion Synthesis \& Editing}
\author {
    Jihoon Kim,\textsuperscript{\rm 1, \rm 2}
    Jiseob Kim, \textsuperscript{\rm 2}
    Sungjoon Choi \textsuperscript{\rm 1}
}
\affiliations {
    \textsuperscript{\rm 1} Korea University\\
    \textsuperscript{\rm 2} Kakao Brain\\
    jihoon-kim@korea.ac.kr, jiseob.kim@kakaobrain.com, sungjoon-choi@korea.ac.kr
}

\begin{document}

\maketitle

\begin{abstract}

Text-based motion generation models are drawing a surge of interest for their potential for automating the motion-making process in the game, animation, or robot industries. In this paper, we propose a diffusion-based motion synthesis and editing model named FLAME. Inspired by the recent successes in diffusion models, we integrate diffusion-based generative models into the motion domain. FLAME can generate high-fidelity motions well aligned with the given text. Also, it can edit the parts of the motion, both frame-wise and joint-wise, without any fine-tuning. FLAME involves a new transformer-based architecture we devise to better handle motion data, which is found to be crucial to manage variable-length motions and well attend to free-form text. In experiments, we show that FLAME achieves state-of-the-art generation performances on three text-motion datasets: HumanML3D, BABEL, and KIT. We also demonstrate that FLAME’s editing capability can be extended to other tasks such as motion prediction or motion in-betweening, which have been previously covered by dedicated models.

\end{abstract}

\section{Introduction}

Given the difficulty of 3D motion generation, automating text-based motion synthesis and editing has been considered a difficult problem despite their usefulness in industries. The process of text-to-motion synthesis and text-based motion editing motion should not only deliver the clear intention of motion but also be able to convey the naturalnesss of human motion at the same time. This effort can be reduced considerably if one can generate motion through language or edit existing motion to desirable motion simply by language. In this study, we present a method that can perform motion synthesis and editing using free-form texts.

\begin{figure}[ht]
\centering

\begin{subfigure}[b]{0.95\columnwidth}
\centering
\includegraphics[width=\columnwidth]{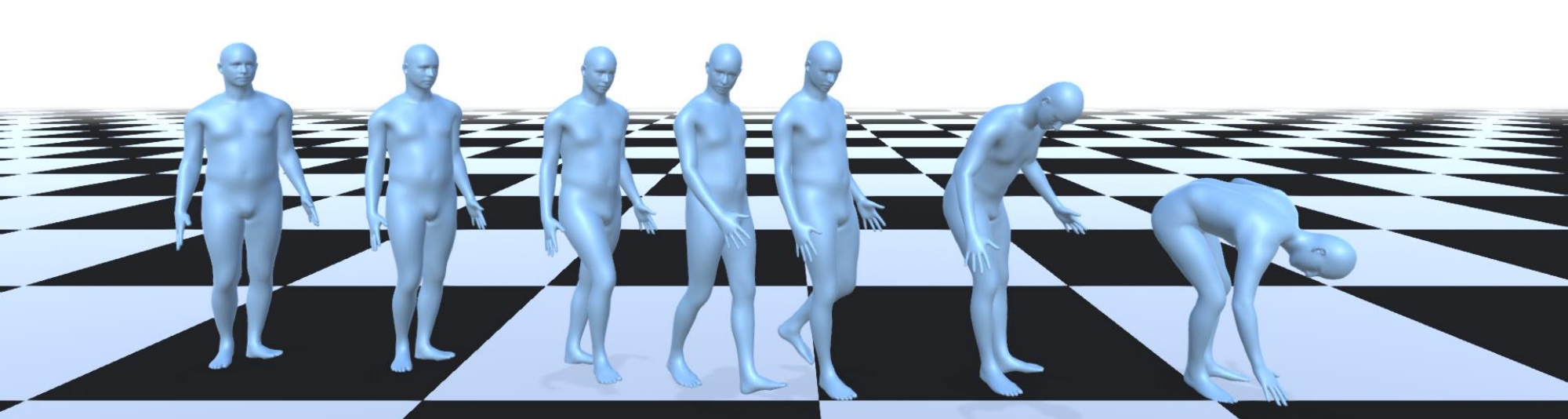} %
\caption{Text-to-motion synthesis result from FLAME with prompt: \textit{``A person walks forward and bends down to pick up something."}}
\end{subfigure}

\begin{subfigure}[b]{0.95\columnwidth}
\centering
\includegraphics[width=\columnwidth]{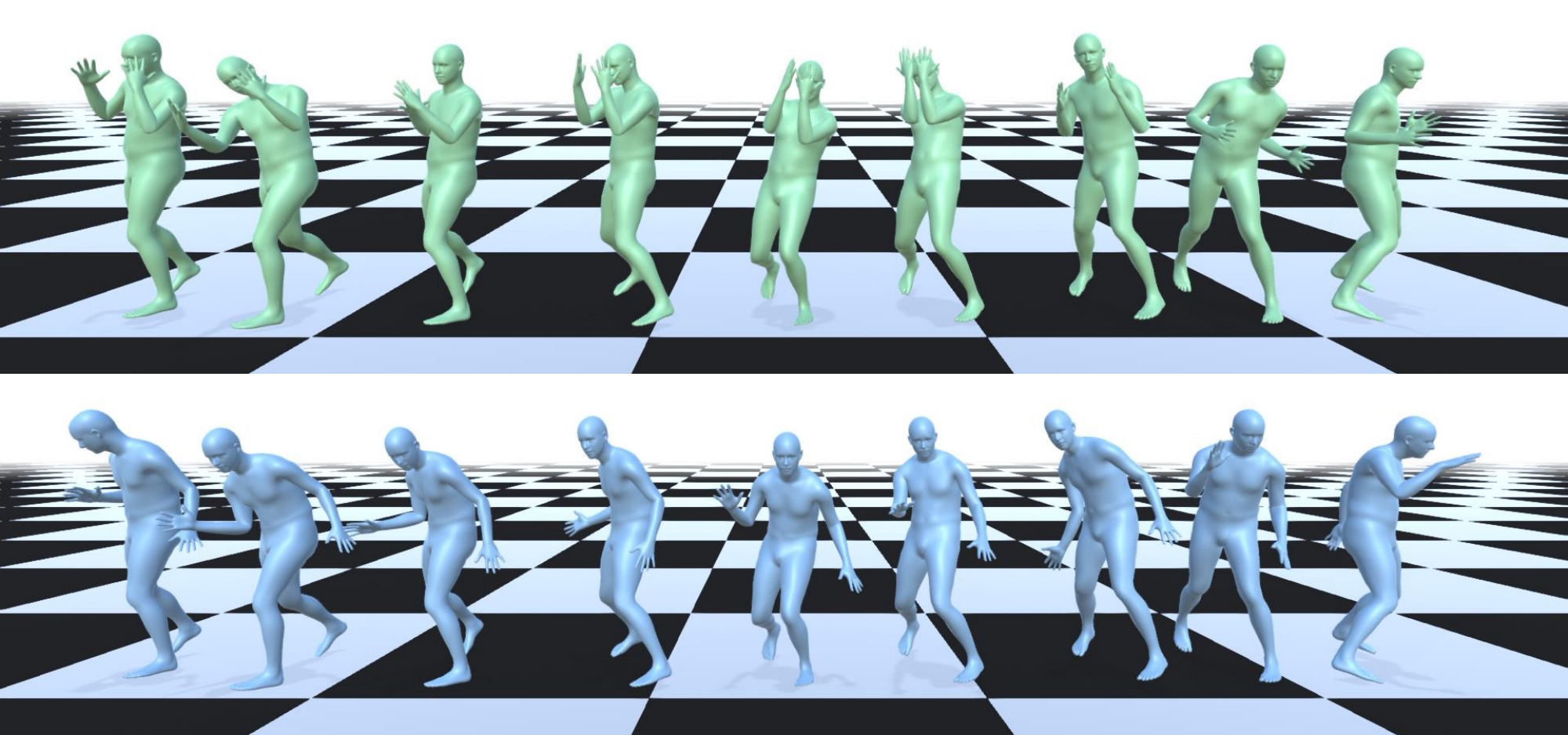}
\caption{(Green) Reference motion. (Blue) Text-based motion editing result from FLAME with prompt: \textit{``A person dribbles a ball."}; The editing model is allowed to edit upper body parts while fixing lower body parts in this example.}
\end{subfigure}

\caption{Overview of text-to-motion synthesis and text-based motion editing. Motion flows from left to right.}
\label{fig:overview}
\end{figure}

Recently, research on generating motion from language has been actively conducted. Many previous studies \citep{guo2020action2motion, petrovich2021action, song2022actformer} have explored methods to synthesize motion from behavioral labels, such as ‘walking’, ‘jumping’, or ‘dancing’ and demonstrated promising results on text-to-motion synthesis tasks. However, synthesizing motion from behavioral labels lacks descriptive power, which limits both diversity and controllability in motion synthesis. As large-scale pre-trained language models (PLMs) advance, there are studies \citep{ghosh2021synthesis, petrovich22temos} take the advantage of using PLMs to generate motion from free-form texts, overcoming the limited expressiveness of simple labels. Although these methods present promising results in synthesizing motion from free-form texts, they lack capability in a flexible conditional generation.

In this paper, we introduce a versatile motion synthesis method that can generate motion well-aligned with the provided prompts and perform editing of a reference motion from textual descriptions. As we aim to present a method that can generate and edit motion, we employ the diffusion model \citep{ho2020denoising, dhariwal2021diffusion, nichol2021glide}, which has been presenting many successful results in image generation and in-painting \citep{nichol2021glide, ramesh2022hierarchical} these days. With this, our proposed method can conduct text-to-motion synthesis and text conditional motion generation, including editing, forecasting, and in-betweening without any fine-tuning or modification of a trained model.

Architectures for diffusion models have been actively studied \citep{ho2020denoising, nichol2021improved, dhariwal2021diffusion} in the image domain, but have not yet been explored in the motion domain. In order to introduce the diffusion model into the motion domain, we focus on the following differences between motion and image. First, unlike the image, which has spatial information without temporal information, motion is inherently spatio-temporal data. Secondly, the length of motion can vary from short motion to long-horizon motion, which requires a model that can handle arbitrary length. To this end, we introduce a transformer decoder-based architecture that can handle temporal aspects and variable lengths. Our proposed method takes diffusion time-step token, motion length token, language tokens, and motion tokens as inputs. Additional language-side information is extracted from PLM and fed to the transformer using cross-attention. Model is trained to learn the denoising process, which gradually reconstructs motion from isotropic Gaussian noise. To sample motion from free-form text, we use classifier-free guidance \cite{ho2021classifier}. We refer to our model as \textbf{FLAME}, which stands for \textbf{F}ree-form \textbf{LA}nguage-based \textbf{M}otion Synthesis and \textbf{E}diting. To the best of our knowledge, \textbf{FLAME} is the first to adopt a diffusion-based generative framework for synthesizing and editing motion.

Our main contributions are summarized as follows:
\begin{itemize}
    \item We propose FLAME, a unified model for motion synthesis and editing with free-form language description.
    \item Our model is the first attempt applying diffusion models to motion data; to handle the temporal nature of motion and variable-length, we devise a new architecture.
    \item We show FLAME can generate more diverse motions corresponding to the same text.
    \item We demonstrate FLAME can solve other classical tasks---prediction and in-betweening---through editing, without any fine-tuning.
\end{itemize}

\section{Related Work}

\subsection{Diffusion Models \& Text-conditional Generation}

Diffusion models \cite{ho2020denoising} are recently proposed generative models that are shown to be good at synthesizing highly-complex image datasets. Compared to GANs \citep{goodfellow2014generative, karras2019style, karras2020analyzing, sauer2022stylegan} and VAEs \citep{kingma2013auto, van2017neural}, they have been presenting improved quality in generating multi-modal outputs, advantageous to text-to-image generation and text-to-motion generation of our interest---there can be various modes of images/motions corresponding to a single text description. 

Diffusion models are originally proposed in \citet{sohl2015deep} and developed in \citet{ho2020denoising} and \citet{song2020denoising}, showing high-quality image generation. After, they are extended to work on conditional generation settings, demonstrating even better performances. Class-conditional models are studied in \citet{dhariwal2021diffusion}, and text-conditional models are proposed by adapting the conditioning scheme for text (GLIDE; \cite{nichol2021glide}). unCLIP\footnote{branded as DALL-E 2 to the public} \cite{ramesh2022hierarchical} and Imagen \cite{saharia2022photorealistic} further show that conditioning on pre-trained high-level embedding gives improved result. Our model shares some similarity with the GLIDE model, but has crucial differences in that it has a new design to handle temporal sequences of variable length.

To achieve the best performance, several techniques have been proposed and applied to the aforementioned models. Improved DDPM \cite{nichol2021improved} suggests learning the reverse-diffusion variances. Classifier-free guidance \cite{ho2021classifier} has been introduced to enable conditional generation without the need for a separate classifier model. We employ both of these techniques in the proposed model.

\begin{figure*}[ht]
\centering
\includegraphics[width=0.95\textwidth]{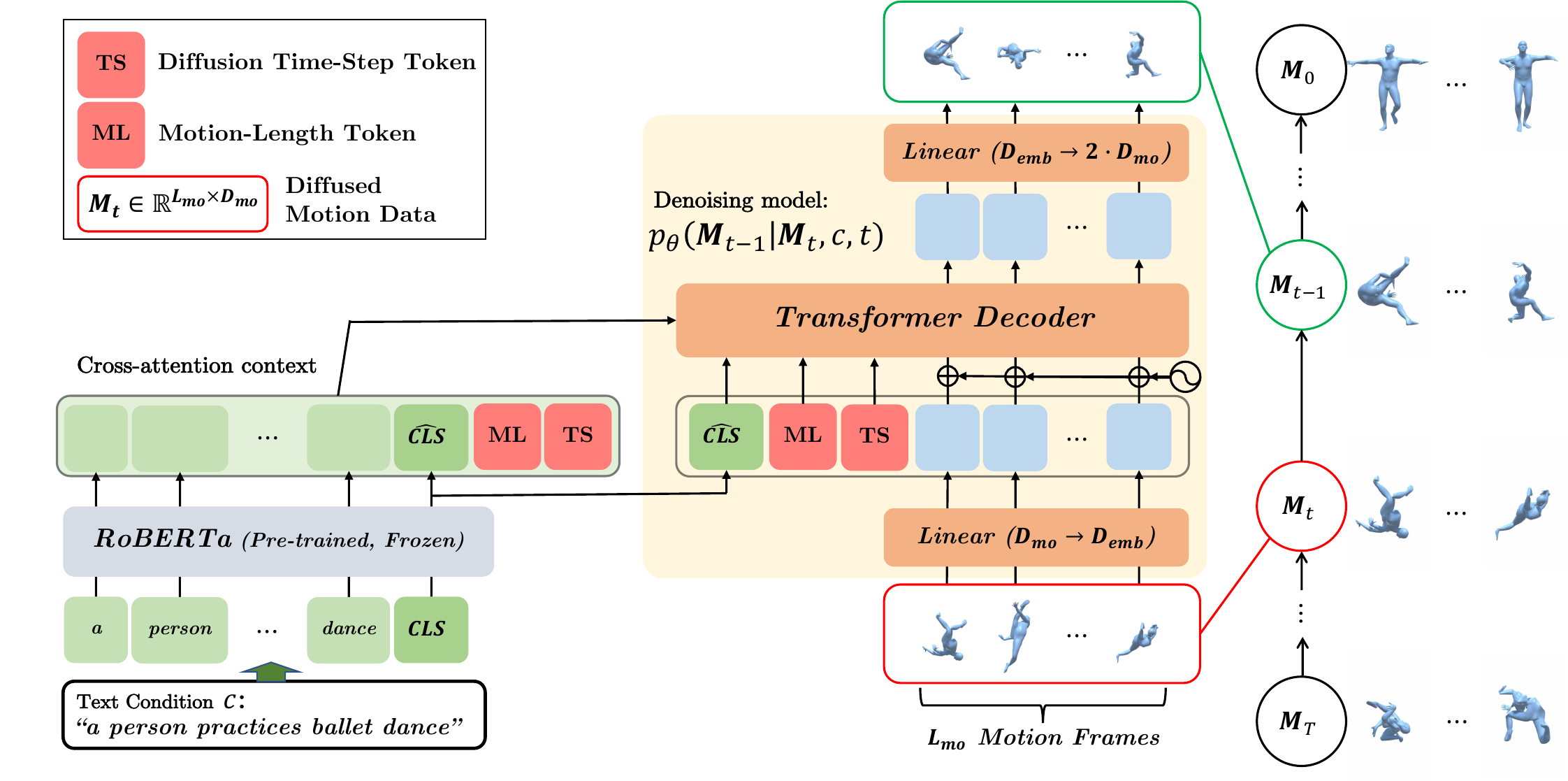}
\caption{\textbf{Overview of the architecture:} FLAME learns the denoising process $p_{\theta}$ from $\boldsymbol{M}_{t}$ to $\boldsymbol{M}_{t-1}$ at diffusion time-step $t$. Input motion is projected and concatenated with language pooler token (CLS), motion-length token (ML), and diffusion time-step token (TS) as input tokens for the transformer decoder. Additional language-side information is fed from a pre-trained frozen language encoder as a cross-attention context. FLAME outputs a $2\cdot D_{mo}$-dimensional sequence of vectors as it predicts both the mean and variance of noise at each diffusion time-steps.}
\label{fig:architecture}
\end{figure*}

\subsection{3D Human Motion Generation}

Motion prediction is a task to forecast subsequent frames from a given frame or multiple frames, and motion in-betweening is a boundary value problem that generates a natural motion sequence while satisfying the given starting and target poses. Motion prediction models \citep{fragkiadaki2015recurrent, guo2019human} and in-betweening models \citep{harvey2020robust, KIM2022108894, Duan_Lin_Zou_Yuan_Qian_Zhang_2022} have been developed, but the models lack the ability to perform multiple tasks with a single model and cannot synthesize motion from textual descriptions.

In the text-to-motion domain, early models \citep{lin:vigil18, ahn2018text2action} approach the text-to-motion synthesis task with the sequence-to-sequence model. After that, \citet{guo2020action2motion} and \citet{petrovich2021action} introduce a variational autoencoder (VAE) to create motion from behavioral labels to improve motion quality and produce a diversified range of motions. Recent models \citep{ghosh2021synthesis, petrovich22temos} advance the text-to-motion task by composing 3D human motion from free-form texts, instead of simple action labels, to cover more expressive human motions. They demonstrate free-form language-based motion generation by taking advantage of pre-trained language model. However, these models have limitations in extensibility to conventional motion tasks or text-based motion editing.

In this study, we propose a model to perform high-quality \textit{text-to-motion synthesis} with flexible editing capability, which can conduct \textit{text-based motion editing} including traditional motion prediction and motion in-betweening without any fine-tuning or modification on a trained generative model.

\section{Proposed Method: FLAME}

We first review the diffusion-based modeling scheme. Then, we explain the model architecture, designed to handle motion data. In the last two subsections, we explain how to do the inference in the synthesis and the editing scenarios using the trained FLAME model.

\subsection{Diffusion-based Model}

The generative modeling scheme of FLAME is inspired by the denoising diffusion probabilistic model (DDPM) \cite{ho2020denoising} and its extension \cite{nichol2021improved}. The general idea of DDPM is to design a diffusion process that gradually adds small amounts of noise to the data and train the model to reverse each of these diffusion steps. The diffusion process eventually converts the data into isotropic Gaussian noise, and thus the fully-trained model would generate samples by repeating denoising steps, starting from pure noise. Essentially, this scheme divides a complex distribution-modeling problem into a set of simple denoising problems.

In details, DDPM defines the diffusion process with the following conditional distribution:
\begin{equation}
    q(\boldsymbol{M}_{t} | \boldsymbol{M}_{t-1})= \mathcal{N}(\boldsymbol{M}_{t}; \sqrt{1-\beta_{t}} \boldsymbol{M}_{t-1}, \beta_{t} \boldsymbol{I}),
\end{equation}
where $\boldsymbol{M}_{t}$ denotes the diffused data at time-step\footnote{Time-steps in this paper denote the diffusion steps. To avoid confusion, the times in motion are deliberately denoted as frames.} $t\in\{0,1,\cdots,T\}$; $\boldsymbol{M}_{0}$ and $\boldsymbol{M}_{T}$ denote the original data and the fully-diffused Gaussian noise, respectively. In case of motion data, $\boldsymbol{M}_{0}$ is the set of all the joint values of the entire frames. $\beta_{t}\in(0,1)$ are hyperparameters with respect to the \textit{variance schedule}, which is set to the \textit{cosine} in FLAME, following \cite{nichol2021improved}. With a sufficiently large $T$ (usually 1,000), this design guarantees $\boldsymbol{M}_{T}\sim \mathcal{N}(\boldsymbol{0},\boldsymbol{I})$. 
Also, it gives the marginal distribution in a closed form: $q(\boldsymbol{M}_{t} | \boldsymbol{M}_{0} )= \mathcal{N}(\boldsymbol{M}_{t}; \sqrt{\bar\alpha_{t}} \boldsymbol{M}_{0}, (1-\bar\alpha_{t})\boldsymbol{I})$, where $\bar\alpha_t=\prod_{i=1}^T \alpha_i$ and $\alpha_t=1-\beta_t$. This allows connecting $\boldsymbol{M}_{t}$ and $\boldsymbol{M}_{0}$ with a single Gaussian noise, which is to be used in formulating Eq.~\ref{eq:loss_simple}.

The model considers the reverse of the diffusion process with the following parameterization:
\begin{equation}
    p_{\theta} (\boldsymbol{M}_{t-1} | \boldsymbol{M}_{t},c) = \mathcal{N}(\boldsymbol{M}_{t-1}; \boldsymbol{\mu}_{\theta}(\boldsymbol{M}_{t}, c, t), \boldsymbol{\Sigma}_{\theta}(\boldsymbol{M}_{t}, c, t)),
\end{equation}
where $c$ is an optional conditioning variable, language description in our problem. Once the model learns this distribution, inference is done by first sampling $\boldsymbol{M}_{T}\sim \mathcal{N}(\boldsymbol{0},\boldsymbol{I})$ and then sampling from $p_{\theta} (\boldsymbol{M}_{t-1} | \boldsymbol{M}_{t},c)$, from $t=T$ to $t=1$.

\paragraph{Training \& Loss Functions}
Training the model parameters, $\theta$, is mostly the same as the VAE \cite{kingma2013auto}. Treating $\boldsymbol{M}_t$ and $\boldsymbol{M}_{t-1}$ as the latent and the data in VAEs, respectively, training is done by maximizing the evidence lower bound (ELBO) for every $t$. In DDPM, it is shown that the core terms in the ELBO loss can be far simplified to the following after a proper re-weighting and reparameterization (see Appendix B for the details):
\begin{equation}
    L_{\textrm{simple}} = \mathbb{E}_{t, \boldsymbol{M}_{0}, \boldsymbol{\epsilon}_t} \left[ \lVert \boldsymbol{\epsilon}_t - \boldsymbol{\epsilon}_{\theta}(\boldsymbol{M}_{t}(\boldsymbol{M}_{0}, \boldsymbol{\epsilon}_t), c, t) \rVert^2 \right],\label{eq:loss_simple}
\end{equation}
where $\boldsymbol{\epsilon}_t \sim \mathcal{N}(\boldsymbol{0},\boldsymbol{I})$ is the noise used to diffuse $\boldsymbol{M}_0$ to make $\boldsymbol{M}_t$. Now, the model is parameterized to sort out the noise component $\boldsymbol{\epsilon}_t$ from $\boldsymbol{M}_t$ instead of considering the mean $\boldsymbol{\mu}_{\theta}$. Performing better in practice, this parameterization still allows computing the mean indirectly from the predicted noise\footnote{$\boldsymbol{\mu}_\theta(\mathbf{M}_t, c, t) = {\frac{1}{\sqrt{\alpha_t}} \Big( \mathbf{M}_t - \frac{\beta_t}{\sqrt{1 - \bar{\alpha}_t}} \mathbf{\epsilon}_\theta(\mathbf{M}_t, c, t) \Big)}$}, and we can sample from $p_{\theta} (\boldsymbol{M}_{t-1} | \boldsymbol{M}_{t},c)$. Note, however, the reverse-process variance $\boldsymbol{\Sigma}_{\theta}$ cannot be learned with this loss as it has been deliberately removed.

To train the variance, \citet{nichol2021improved} proposes a hybrid loss and demonstrates better performance:
\begin{equation}
    L_{\textrm{hybrid}} = L_{\textrm{simple}} + \lambda L_{\textrm{vlb}},
\end{equation}
where $L_{\textrm{vlb}}$ is the original ELBO loss without the re-weighting (see Appendix B for the details). We also use this loss for training FLAME.


\subsection{Model Architecture for Motion Data}
Unlike images, motion involves temporal as well as spatial patterns, and its length varies by samples. Thus, we cannot use the U-Net-based architectures widely adopted in the previous diffusion models; we instead propose a new transformer-based architecture (see Figure \ref{fig:architecture}).


\subsubsection{Transformer Decoder}
As explained in the previous section, the model learns the denoising distribution, $p_{\theta} (\boldsymbol{M}_{t-1} | \boldsymbol{M}_{t},c)$, and we use transformer decoder to implement this. As an input to the transformer, the diffused motion $\boldsymbol{M}_t$ is presented as a sequence of $L_{mo}$ frames, where each frame consists of $D_{mo}$-dimensional joint angle values. To be used as tokens, each frame passes through a linear layer, converted to be $D_{emb}$-dimensional (see Fig.~\ref{fig:architecture}). The conditioning with a language description $c$ is implemented as a cross-attention context. The context is a sequence of token embeddings computed from a pre-trained language model (to be explained in the next paragraph). The output frames are collected at positions where the motion tokens are processed. Then they pass through a linear layer, but this time converted to $2 \cdot D_{emb}$-dimensional vectors to learn both mean and variance. The vectors are concatenations of $\epsilon$ and $\Sigma$, which fully parameterize $p_{\theta} (\boldsymbol{M}_{t-1} | \boldsymbol{M}_{t},c)$ together. Variable length is handled by masking in the transformer decoder.

\subsubsection{Pre-trained Language Model (PLM)}
We use the pre-trained RoBERTa model \cite{liu2019roberta} to encode the textual description into a sequence of high-level token embeddings. PLMs have been showing their language understanding capability can be transferred to a variety of tasks. Hence, we also adopt the PLM to extract language features.

\subsubsection{Time-Step (TS) and Motion-Length (ML) Tokens}
In addition to the language and the motion tokens of length $L_{lang}$ and $L_{mo}$, respectively, we introduce two special embedding tokens as inputs to the transformer. The time-step token (TS) is to give the time-step information $t$, and the motion-length token (ML) is to give the motion-length information $L_{mo}$ to the model. Since FLAME generates the entire motion at once, unlike the autoregressive generation using causal masks, these tokens can explicitly inform the network about motions to be generated.

\begin{figure}[ht]
\centering
\includegraphics[width=0.95\columnwidth]{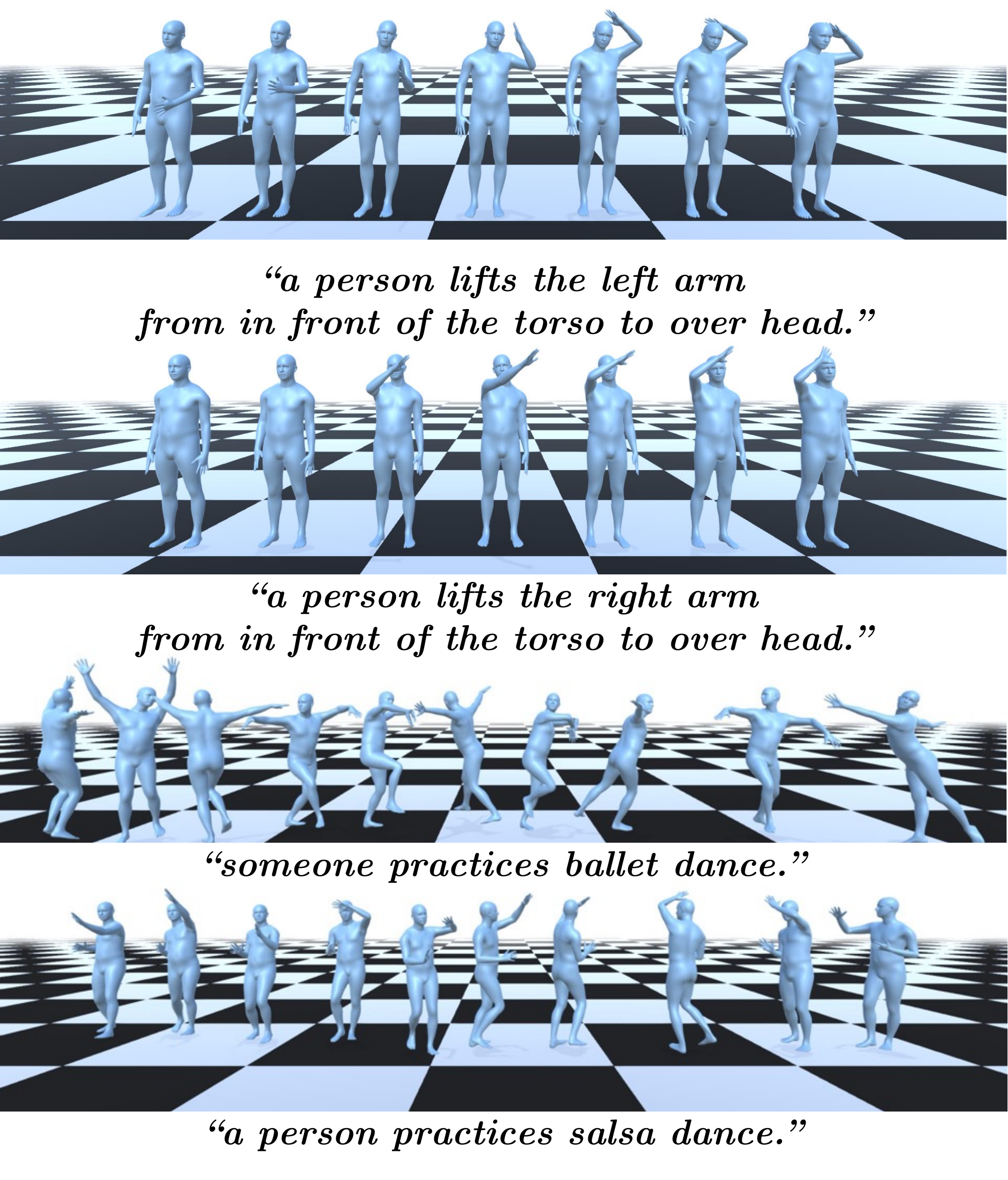}





\caption{Qualitative results on text-to-motion synthesis task. FLAME is able to synthesize motion from detailed textual descriptions. Motion sequences flow from left to right.}
\label{fig:text-to-motion}
\end{figure}

\subsection{Inference for Motion Synthesis}
When synthesizing motion from text, we use the classifier-free guidance \cite{ho2021classifier} technique for better semantic alignment. While the guidance trades off the sample diversity by little, it uplifts the precision by large and is used in many text-to-image generation models \citep{nichol2021glide, ramesh2022hierarchical}. 

In details, the guidance amplifies the effect of the conditioning variable $c$ when predicting the noise:
\begin{equation}
    \hat{\epsilon}_{\theta} (\mathbf{M}_{t} \mid c) = \epsilon_{\theta}(\mathbf{M}_{t} \mid \emptyset) + s \cdot \left( \epsilon_{\theta} (\mathbf{M}_{t} \mid c) - \epsilon_{\theta} (\mathbf{M}_{t} \mid \emptyset ) \right).
\end{equation}
At each denoising step, a guided version $\hat{\epsilon}_{\theta} (\mathbf{M}_{t} \mid c)$ is used instead of the original prediction $\epsilon_{\theta} (\mathbf{M}_{t} \mid c)$, which amplifies the conditioning effect by a scalar amount $s>1$. To get an unconditioned prediction as well from the model, $\epsilon_{\theta}(\mathbf{M}_{t} \mid \emptyset)$, we randomly replace the text with an empty string $\emptyset$ during training.

\begin{table*}[ht]
\centering
\resizebox{0.95\textwidth}{!}{%
\begin{tabular}{c|cccccc|cccccc}
\hline
                  & \multicolumn{6}{c|}{HumanML3D}                                         & \multicolumn{6}{c}{BABEL}                                            \\ \cline{2-13} 
Method            &       &         &        & \multicolumn{3}{c|}{R-Precision $\uparrow$} &       &        &        & \multicolumn{3}{c}{R-Precision $\uparrow$} \\ \cline{5-7} \cline{11-13} 
 &
  mCLIP $\uparrow$ &
  FD $\downarrow$ &
  MID $\uparrow$ &
  Top-1 &
  Top-2 &
  Top-3 &
  mCLIP $\uparrow$ &
  FD $\downarrow$ &
  MID $\uparrow$ &
  Top-1 &
  Top-2 &
  Top-3 \\ \hline
\citet{lin:vigil18}   & 0.142 & 58.694  & 18.141 & 0.225         & 0.298        & 0.363        & 0.183 & 51.873 & 33.967 & 0.678        & 0.709        & 0.754        \\
Language2Pose     & 0.145 & 55.365  & 18.982 & 0.233         & 0.305        & 0.381        & 0.199 & 42.209 & 39.360 & 0.685        & 0.713        & 0.788        \\
\citet{ghosh2021synthesis} & 0.106 & 109.778 & 14.643 & 0.122         & 0.151        & 0.204        & 0.150 & 75.316 & 28.363 & 0.505        & 0.591        & 0.653        \\
TEMOS             & 0.254 & 49.142  & 28.570 & 0.355         & 0.481        & 0.589        & 0.273 & 38.679 & 46.953 & 0.786        & 0.835        & 0.893        \\
Guo et al. (2022) & 0.281 & 27.950 & 27.744 & 0.452 & 0.611 & 0.675 & 0.301 & 24.882 & 44.758 & 0.832 & 0.894 & 0.911 \\
FLAME (Ours) &
  \textbf{0.297} &
  \textbf{21.152} &
  \textbf{29.935} &
  \textbf{0.513} &
  \textbf{0.673} &
  \textbf{0.749} &
  \textbf{0.318} &
  \textbf{18.234} &
  \textbf{53.003} &
  \textbf{0.888} &
  \textbf{0.926} &
  \textbf{0.939} \\ \hline
\end{tabular}%
}
\caption{Text-to-motion benchmark on the HumanML3D and BABEL.}
\label{tab:results-hml3d-babel}
\end{table*}



\begin{table*}[ht]
\centering
\resizebox{0.95\textwidth}{!}{%
\begin{tabular}{c|cccc|cccc}
\hline
& \multicolumn{4}{c|}{Average Positional Error $\downarrow$} & \multicolumn{4}{c}{Average Variance Error $\downarrow$} \\ \cline{2-9} 
Method & root joint & global traj & mean local & mean global & root joint & global traj & mean local & mean global \\ \hline
Lin et al. (2018) & 1.966 & 1.956 & 0.105 & 1.969 & 0.790 & 0.789 & 0.007 & 0.791 \\
Language2Pose & 1.622 & 1.616 & \textbf{0.097} & 1.630  & 0.669 & 0.669 & 0.006 & 0.672 \\
Ghosh et al. (2021) & 1.291 & 1.242 & 0.206 & 1.294 & 0.564 & 0.548 & 0.024 & 0.563 \\
TEMOS & 0.963 & 0.955 & 0.104 & 0.976 & \textbf{0.445} & \textbf{0.445} & \textbf{0.005} & \textbf{0.448} \\
Guo et al. (2022) & 0.949 & 0.937 & 0.108 & 0.940 & 0.510 & 0.507 & 0.007 & 0.552 \\
FLAME (Ours) & \textbf{0.881} & \textbf{0.869} & 0.110 & \textbf{0.899} & 0.497 & 0.495 & 0.007 & 0.500 \\ \hline
\end{tabular}%
}
\caption{APE and AVE benchmark on the KIT dataset.}
\label{tab:results-kit}
\end{table*}

\begin{table}[ht]
\centering
\resizebox{\columnwidth}{!}{%
\begin{tabular}{c|ccc}
\hline & mCLIP$\uparrow$ & Joint Variance $\uparrow$ & Multimodality $\uparrow$  \\ \hline
TEMOS & 0.252 & 0.017 & 11.901 \\
FLAME (Ours) & 0.298 & 0.072 & 31.500 \\ \hline
\end{tabular}%
}
\caption{Diversity evaluation on HumanML3D. Each model generates 10 samples per text in the test set for this evaluation.}
\label{tab:diversity}
\end{table}

\subsection{Inference for Motion Editing}
In motion editing, we want to manipulate parts of data, either frame-wise, joint-wise, or both. Similarly to image in-painting \cite{banitalebi2021repaint}, we take a ``diffuse then conditionally denoise" strategy to in-fill the editable parts with the given language condition. This way, we can make a bridge between the unedited data and the edited data distributions.

In details, we are given with data $\boldsymbol{M}_0^{\text{ref}}$ to edit and a binary mask $m$ that designates the parts for editing with zeros, and ones otherwise. We first diffuse $\boldsymbol{M}_0^{\text{ref}}$ in the same way as done in training, obtaining $\boldsymbol{M}_t^{\text{ref}}$ for every $t$. Then, the fully diffused data $\boldsymbol{M}_T^{\text{ref}}$ is denoised step-by-step using the trained model; however, the masked parts (where $m=1$) are now overwritten with the unedited ground truth, or the reference $\boldsymbol{M}_{t-1}^{\text{ref}}$:
\begin{equation}
    \boldsymbol{M}_{t-1}^{\text{edit}} = (1-m) \odot \boldsymbol{M}_{t-1}^{\text{pred}} + m \odot \boldsymbol{M}_{t-1}^{\text{ref}}.
\end{equation}
Here, $\boldsymbol{M}_{t-1}^{\text{pred}} \sim p_{\theta} (\boldsymbol{M}_{t-1} | \boldsymbol{M}_{t}^{\text{edit}},c^{\text{edit}})$ is a predicted denoised sample by the model with a new condition $c^{\text{edit}}$ (note $\boldsymbol{M}_T^{\text{edit}} := \boldsymbol{M}_T^{\text{ref}}$). 

\section{Experiments}

\begin{table*}[ht]
    \centering
    \begin{tabular}{cccc|cccccc}
    \hline
    & & & & & & & \multicolumn{3}{c}{R-Precision $\uparrow$} \\ \cline{8-10} 
    Self-Attn & ML Token & X-Attn & PLM Freeze & mCLIP $\uparrow$ & FD $\downarrow$ & MID $\uparrow$ & Top-1 & Top-2 & Top-3 \\ \hline
    \checkmark &  &  &  & 0.239 & 60.142 & 20.980 & 0.405 & 0.445 & 0.460 \\
    \checkmark & \checkmark & & & 0.239 & 59.254 & 21.301 & 0.410 & 0.441 & 0.465 \\
    \checkmark & \checkmark & \checkmark & & 0.290 & 27.157 & 29.010 & 0.439 & 0.580 & 0.654 \\
    \checkmark & \checkmark & \checkmark & \checkmark & 0.297 & 21.152 & 29.935 & 0.513 & 0.673 & 0.749 \\ \hline
    \end{tabular}%
    \caption{Ablation study on four components of FLAME on the HumanML3D.}
    \label{tab:ablation-study}
\end{table*}

\subsection{Datasets}

\begin{figure}[ht]
    \centering
    \includegraphics[width=0.95\columnwidth]{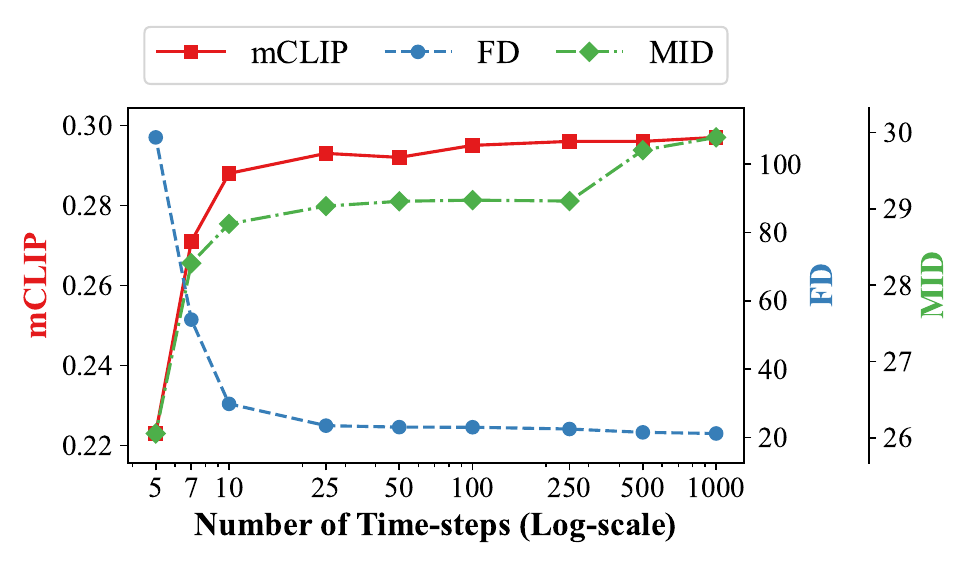}
    \caption{Quantitative results with different numbers of reduced sampling steps. The same trained model with $T=1000$ diffusion time-steps is used.}
    \label{fig:diffusion-sampling-steps-ddpm}
\end{figure}

In the experiments, we train and evaluate our model on the following datasets. Detailed preprocessing is described in Appendix A.

\begin{enumerate}
    \item \textbf{HumanML3D\textsubscript{SMPL} \cite{guo2022generating}} is a recently proposed large motion-text pair dataset containing 44,970 full-sentence text descriptions for 14,616 motions from AMASS \cite{AMASS:ICCV:2019} and HumanAct12 \cite{guo2020action2motion}. We use SMPL motion data from AMASS directly for the annotation set.
    \item \textbf{BABEL \cite{BABEL:CVPR:2021}} provides a language description for AMASS. We use 63,353 frame-level annotations to precisely represent the semantics of motion.
    \item \textbf{KIT \cite{plappert2016kit}} consists of 3,911 motion sequences paired with 6,353 textual descriptions. We follow the evaluation protocol used by TEMOS \cite{petrovich22temos}.
\end{enumerate}

\begin{table}[ht]
\centering
\resizebox{\columnwidth}{!}{%
\begin{tabular}{c|cccccccc}
\hline
Sampling Steps   & 5     & 10    & 25    & 50    & 100   & 250   & 500   & 1,000  \\ \hline
Time Elapsed (s) & 0.72 & 0.84 & 1.31 & 2.17 & 3.70 & 8.61 & 16.66 & 32.81 \\ \hline
\end{tabular}%
}
\caption{Elapsed time for sampling a motion. Performance is recorded on a single NVIDIA's Tesla V100 SXM2 32GB machine.}
\label{tab:time-elapsed}
\end{table}

\subsection{Motion Representation}


\subsubsection{HumanML3D\textsubscript{SMPL} and BABEL}
To represent motion, we use the coordinates of the root joint $\boldsymbol{r}_{\text{root}} \in \mathbb{R}^{3}$ and the rotations of 24 SMPL-joints \cite{SMPL:2015} with respect to their parent joints. We use the SMPL pose parameters directly, instead of employing a customized skeleton for simplicity and compatibility. We adopt 6D representation \cite{zhou2019continuity} to describe rotations rather than the axis-angle format. In total, a single pose $\boldsymbol{p}$ is represented with a 147-dimensional vector $\boldsymbol{p} \in \mathbb{R}^{147=3+24\times6}$, and motion is represented with a sequence of pose vectors $\boldsymbol{M} = [\boldsymbol{p}_{1}, \boldsymbol{p}_{2}, \cdots, \boldsymbol{p}_{L_{mo}}] \in \mathbb{R}^{L_{mo} \times 147}$.

\subsubsection{KIT}
For consistency with the prior work, we follow the motion representation used by TEMOS on the KIT dataset. The human pose is encoded with a 64-dimensional feature vector $\boldsymbol{p} \in \mathbb{R}^{64}$ composed of coordinates for 20 joints, an angle between the local and global coordinate system, and translation.

\begin{figure}[ht]
\captionsetup[subfigure]{labelformat=empty}
\centering
\includegraphics[width=0.95\columnwidth]{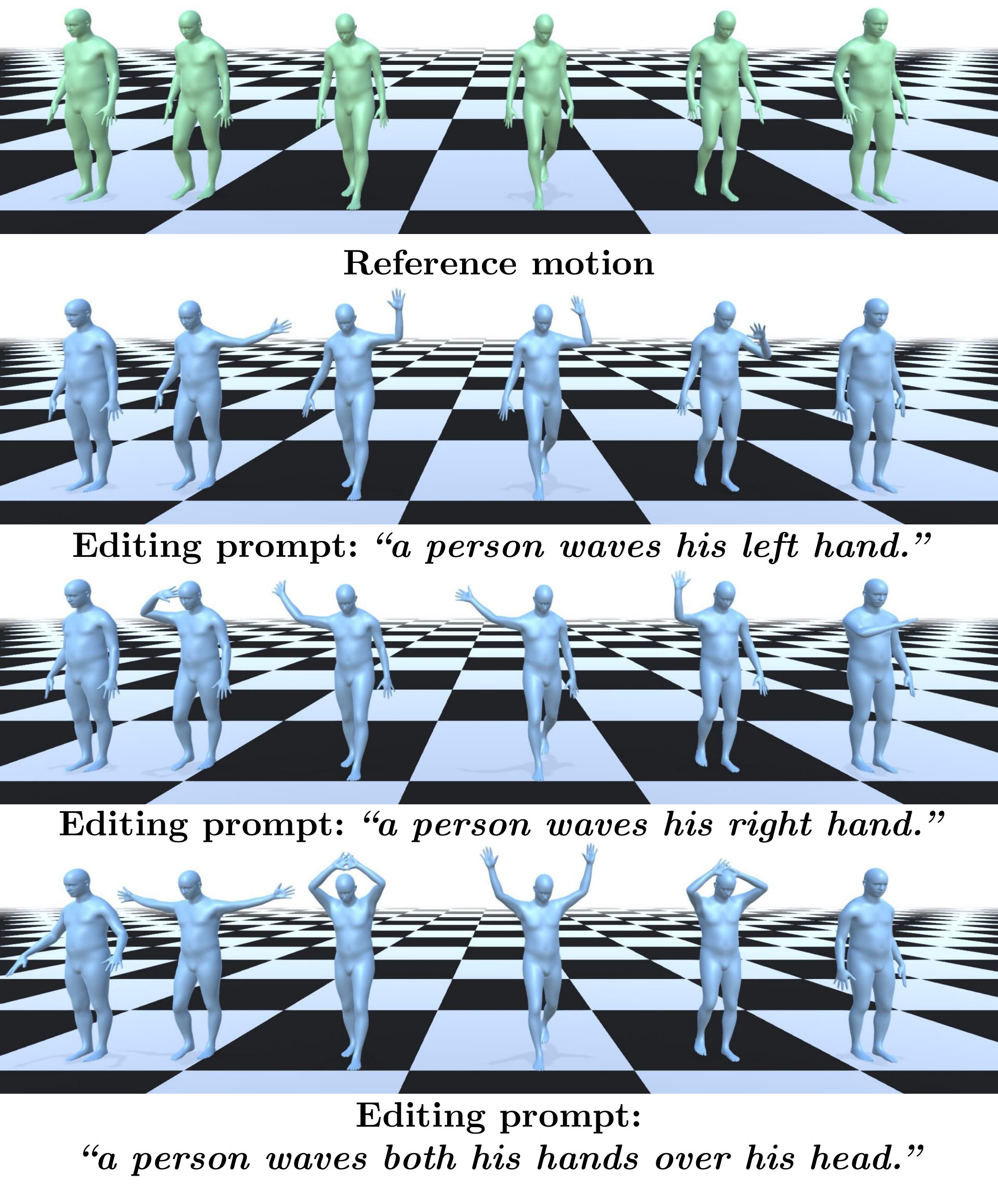}




\caption{Qualitative results on text-based motion editing. FLAME edits reference motion with given prompts. The model is allowed to edit from both shoulders to hands in this motion. Motion flows from left to right.}
\label{fig:edit-motion}
\end{figure}

\begin{figure}[ht]
\captionsetup[subfigure]{labelformat=empty}
\centering

\begin{subfigure}[b]{0.95\columnwidth}
\centering
\includegraphics[width=\columnwidth]{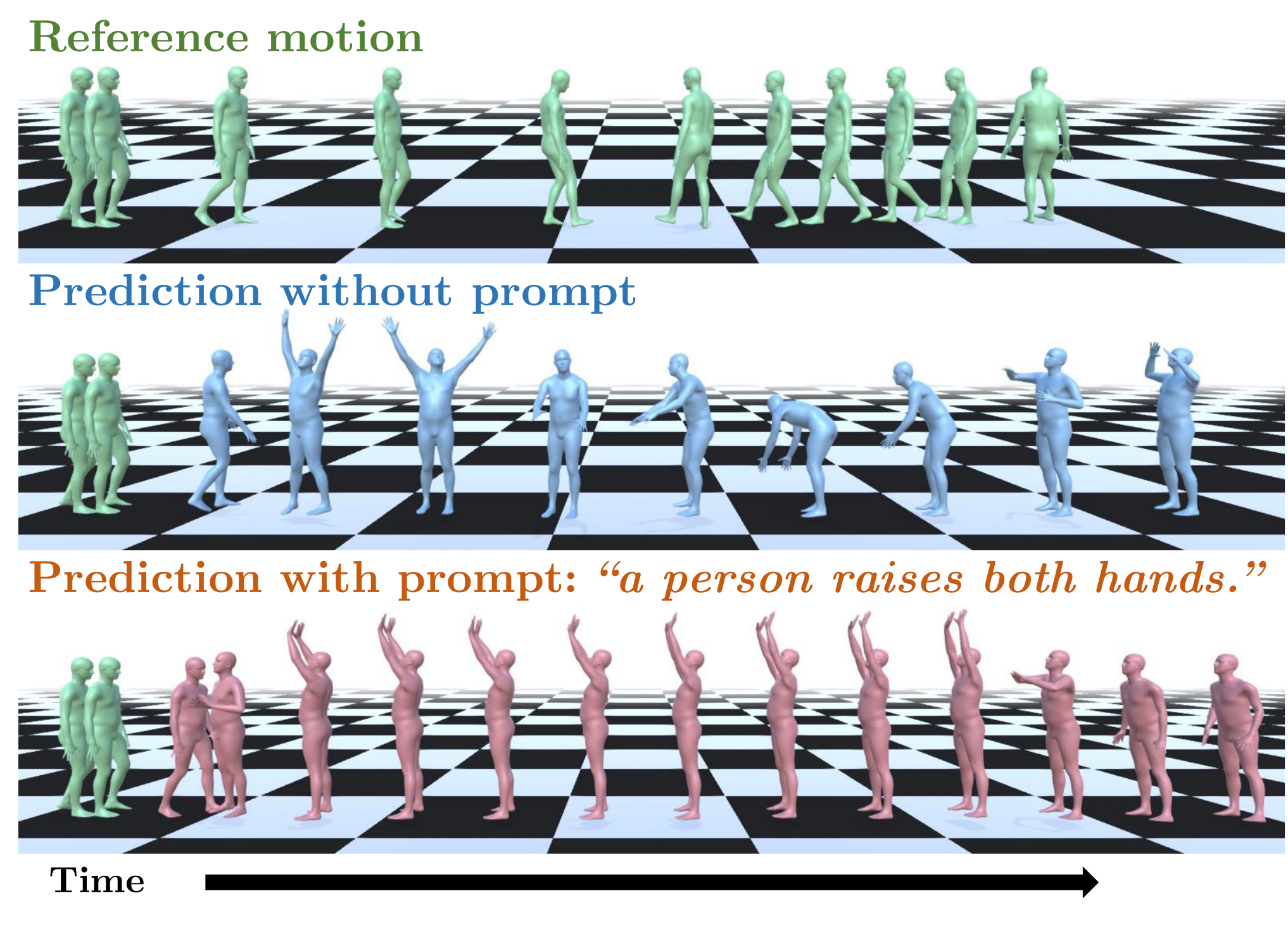} %
\label{fig:motion-prediction}
\end{subfigure}

\begin{subfigure}[b]{0.95\columnwidth}
\centering
\includegraphics[width=\columnwidth]{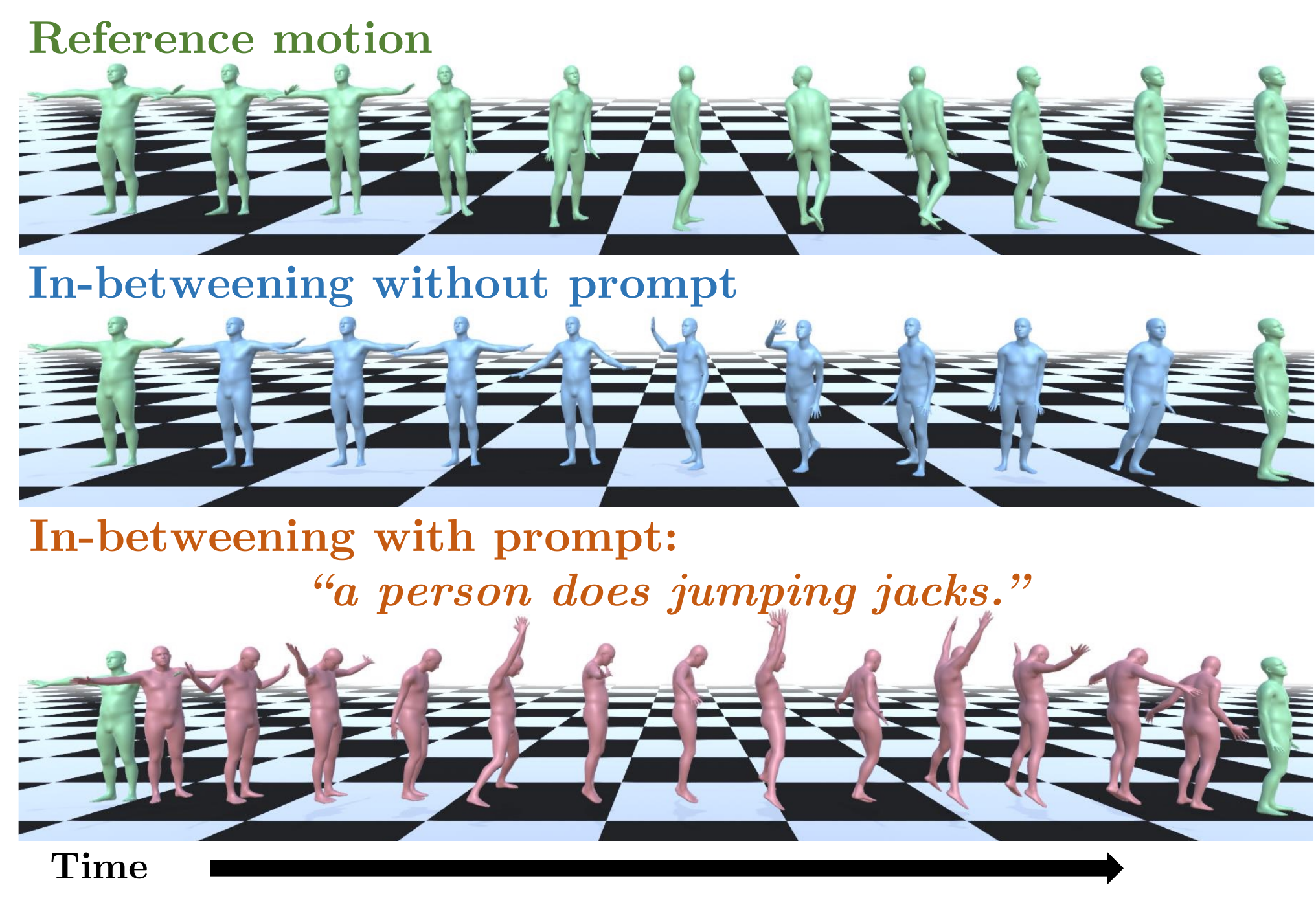} %
\end{subfigure}

\caption{Application of FLAME on motion prediction and motion in-betweening. Green poses are conditioning frames.}
\label{fig:motion-prediction-and-in-betweening}
\end{figure}

\subsection{Evaluation Metrics}

\subsubsection{APE and AVE}
The Average Position Error (APE) measures the mean positional difference for a generated motion against the ground-truth motion, and the Average Variance Error (AVE) measures the difference of variances between the generated and ground-truth motion. \citet{ahuja2019language2pose}, \citet{ghosh2021synthesis}, and \citet{petrovich22temos} used the APE and AVE as quantitative metrics for text-to-motion task evaluation on the KIT dataset (see Appendix C for the details).

\subsubsection{Feature Extractor}
Although APE and AVE are used in the previous work, the metrics have a limitation in that they only rely on the joint values of the reference motion instead of high-level semantics. This problem has been complemented by using CLIP \cite{radford2021learning} score or FID in image-text domain. In a similar vein, we separately train a motion and text encoder in a contrastive manner using InfoNCE loss \cite{oord2018representation}. This model is used to compute motion-text alignment (mCLIP), Fréchet distance (FD), mutual information divergence (MID) \cite{KIM2022108894}, and R-Precision.

\subsubsection{Motion CLIP Score (mCLIP)} Motion CLIP score (mCLIP) computes motion-text alignment by computing the cosine similarity between motion and text embeddings from the separately trained motion CLIP model and denote the similarity as \textit{mCLIP}.

\subsubsection{Fréchet Distance (FD)} FID \cite{heusel2017gans} has been used in the image generation domain combined with the Inception model \cite{szegedy2016rethinking} to measure the distance between the real and generated image feature vectors. We use the same concept in this work by replacing image feature vectors with motion feature vectors.

\subsubsection{Mutual Information Divergence (MID)} Similar to mCLIP, the metric measures the alignment between different modalities, but it measures the alignment based on cross-mutual information instead of cosine similarity. MID \cite{kim2022mutual} is recently proposed as a unified metric to evaluate multimodal generative models.

\subsubsection{R-Precision} R-precision is a metric to measure the alignment between the generated motion and prompt, proposed by \citet{guo2022generating}. During sampling, it generates 32 textual descriptions composed of one ground truth and randomly sampled 31 texts from the test annotation pool. R-precision counts the average retrieval accuracy by ranking the motion feature and text feature by Euclidean distance.

\subsection{Training Details}
Our FLAME model uses 1,000 diffusion time steps to learn the reverse process with cosine beta scheduling \cite{nichol2021improved}. AdamW \cite{loshchilov2017decoupled} is used for experiments with learning rate of 0.0001 and weight decay of 0.0001. For classifier-free guidance, 25\% of texts are replaced with empty strings during training, and the classifier guidance scale of 8.0 is used for sampling. FLAME is backed by 8 transformer decoder layers with 8 heads, 2,048 feedforward dimensions, and 768 embedding dimensions ($D_{emb}$) for both motion and text features (total 64M trainable parameters, see Appendix D for the details). We train the FLAME model using 4 $\times$ NVIDIA Tesla V100 SXM2 32GB for 600K steps on the HumanML3D, 1M steps on the BABEL, and 200K steps on the KIT dataset.


\subsection{Quantitative Results on Text-to-motion}
We compare our method to four state-of-the-art models: \citet{lin:vigil18}, Language2Pose \cite{ahuja2019language2pose}, \citet{ghosh2021synthesis}, TEMOS \cite{petrovich22temos}, and \citet{guo2022generating}. In case of comparing models using PLM, we replace the PLM with the same model, RoBERTa, to prevent the selection of PLM from influencing the benchmark. Table~\ref{tab:results-hml3d-babel} and Table~\ref{tab:results-kit} present the benchmark results on the three datasets. For fair comparison on KIT dataset, we evaluate our model on the same pipeline used in TEMOS. FLAME outperforms other models on all metrics except for the variance metrics in Table~\ref{tab:results-kit}. However, large variance in motion does not necessarily mean low quality, for example, a prompt ``a person dribbles a ball." can correspond to a diverse set of motions rather than a single corresponding ground-truth motion. To validate this, we further compare generated motions on three metrics. First, we sample 10 motions per text annotation in the test set, then we average the variance of joints for the 10 generated motions (Joint Variance). Multimodality of motion (\citet{guo2020action2motion}) is employed to measure the diversity of generated motions. We also compute the average mCLIP score to support that generated motions are not only diverse but also well-aligned to the prompt. Table~\ref{tab:diversity} summarizes the results. All reported metrics are averaged after three trials.

\subsection{Ablation Study}
An ablation study is conducted to validate the four components in our proposed architecture: self-attention block, motion length token, cross-attention block, and freezing of the language model. We start our model from transformer encoder architecture, which uses self-attention only. Next, we add a motion length token to input tokens to explicitly feed the model the number of frames to be generated. On top of these, we employ the cross-attention mechanism, using the transformer decoder architecture. To make the cross-attention context more expressive, we include the first 20 tokens from the PLM output along with the CLS token output. Lastly, we freeze the PLM during the training stage, which results in considerable improvement in performance. These are provided in Table~\ref{tab:ablation-study}.

One of the major drawbacks of diffusion-based models is the slow sampling speed. As provided in Table~\ref{tab:time-elapsed}, sampling using 1,000 diffusion steps takes more than 30 seconds per sample, which hinders its use in practical application. To improve sampling speed, we reduced sampling steps from the same trained model and empirically observed the reduced sampling steps can maintain sample quality unless the diffusion step is extremely reduced (Figure~\ref{fig:diffusion-sampling-steps-ddpm}).

\subsection{Application on Other Motion Tasks}
The proposed motion editing method can be extended to other motion tasks: motion prediction and in-betweening. Unlike most previous task-specific methods, FLAME can perform various motion tasks due to its flexible conditional generation capability (Figure~\ref{fig:motion-prediction-and-in-betweening}).

\section{Conclusion}
In this study, we explored a unified model to perform text-to-motion generation and text-based motion editing. To achieve the objective, we proposed a diffusion-based motion generative model FLAME, which is distinguished from previous work in terms of sample quality and flexibility in conditional generations. We expect our proposed model can greatly streamline the laborious motion generation process and lower the barrier to 3D motion synthesis. In the future, we would like to improve the sampling strategy to enable real-time application and make use of features learned in other domains such as the image-vision domain.

\section{Acknowledgments}
This work was supported by Institute of Information \& communications Technology Planning \& Evaluation (IITP) grant funded by the Korea government (MSIT) (No. 2019-0-00079 and No. 2022-0-00612).

\bibliography{aaai23}
\end{document}


\maketitle
\appendix

\section{Appendix A: Data Preprocessing}

\subsection{HumanML3D}
HumanML3D \cite{guo2022generating} is an annotation dataset for the two datasets: AMASS \cite{AMASS:ICCV:2019} and HumanAct12 \cite{guo2020action2motion}. In AMASS, motions are represented in SMPL format, which includes `gender', `framerates', `translation', and `pose parameters'. Although the SMPL format can represent motion including finger or facial expression, the majority of the motion data in the dataset only contain information about 24 body joint values, so we use the 24 joints to represent motion. Motions in HumanAct12 are transformed to SMPL format using the pipeline introduced from \cite{petrovich2021action}. We follow the original HumanML3D protocol \cite{guo2022generating} to split train, validation, and test set. Motions are processed at 20Hz.

\subsection{BABEL}
Annotations in BABEL \cite{BABEL:CVPR:2021} are categorized into two types: sequence labels and frame labels. SequenceSequence labels describe the entire action in a sequence, while a frame label depicts relatively short motion clips. Therefore, frame labels are more precisely aligned to motion. For this reason, we only use frame-level annotation in experiments. Same as the HumanML3D preprocessing, we use the SMPL representation directly. The BABEL dataset is split into 60\% of training data, 20\% of validation data, and 20\% of test data. However, the test set is not publicly accessible since it is set aside for challenge. Hence, we split 20\% of the validation set as a test set for benchmarks. The motion lengths of BABEL's frame labels are distributed from nearly zero to more than two minutes. (Raw BABEL frame-level annotation metadata contains some clear mislabeling such as the annotation starting time located behind the annotation end time or missing time information. We exclude such cases in preprocessing pipeline.) We preprocess all motion clips to have 128 frames (20Hz), which can cover about 90\% of the entire dataset, to facilitate the model to learn text-to-motion synthesis tasks. For the sequence duration greater than 128 frames, we use a sliding window with offset frames (10 frames) to clip the motion to have predefined frames. For the motion sequence shorter than 128 frames, the duration of the motion clip is increased to 128 frames by extending the annotation end time. Annotation is unchanged for both cases. Motions are processed at 20Hz.

\begin{figure}[h]
\centering

\begin{subfigure}[t]{0.45\columnwidth}
\centering
\includegraphics[width=\columnwidth]{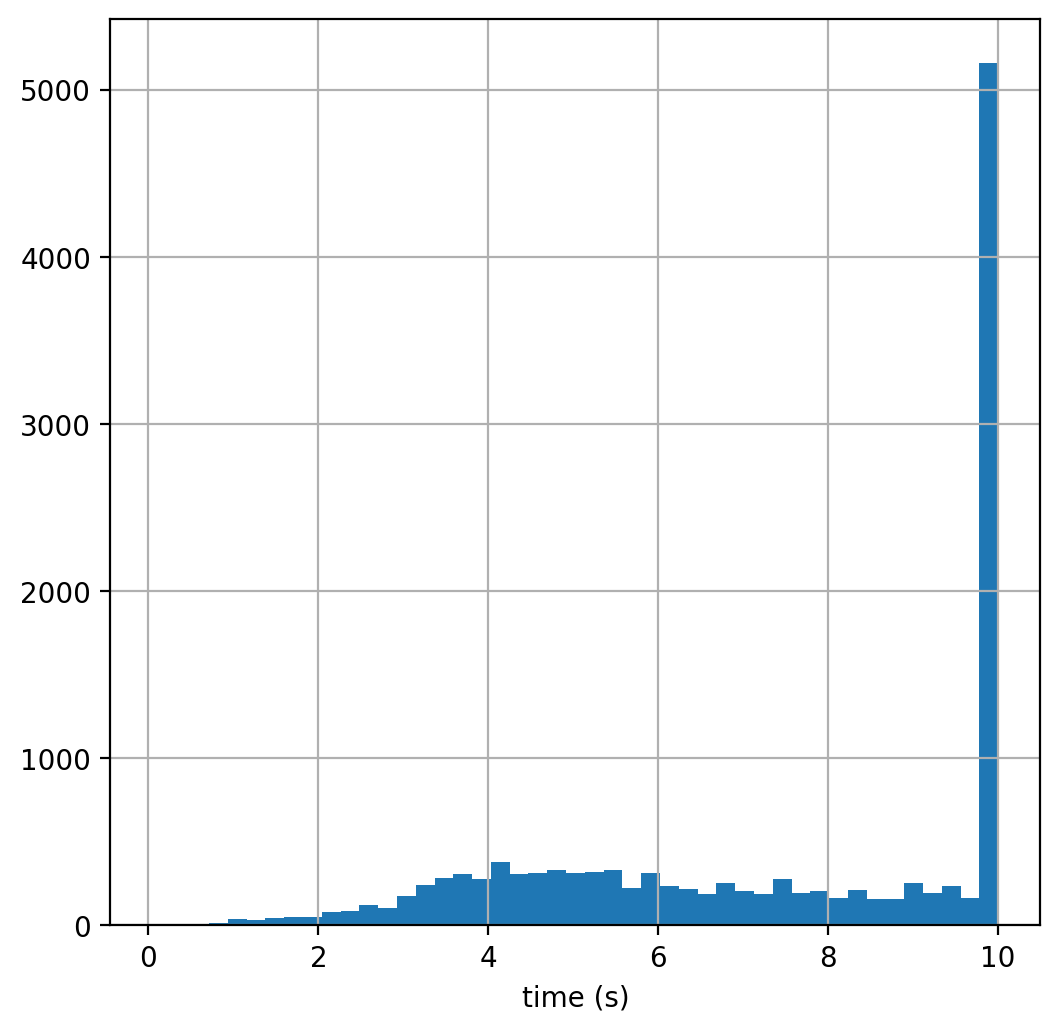} 
\caption{HumanML3D}
\label{fig:motion-prediction}
\end{subfigure}%
~
\begin{subfigure}[t]{0.45\columnwidth}
\centering
\includegraphics[width=\columnwidth]{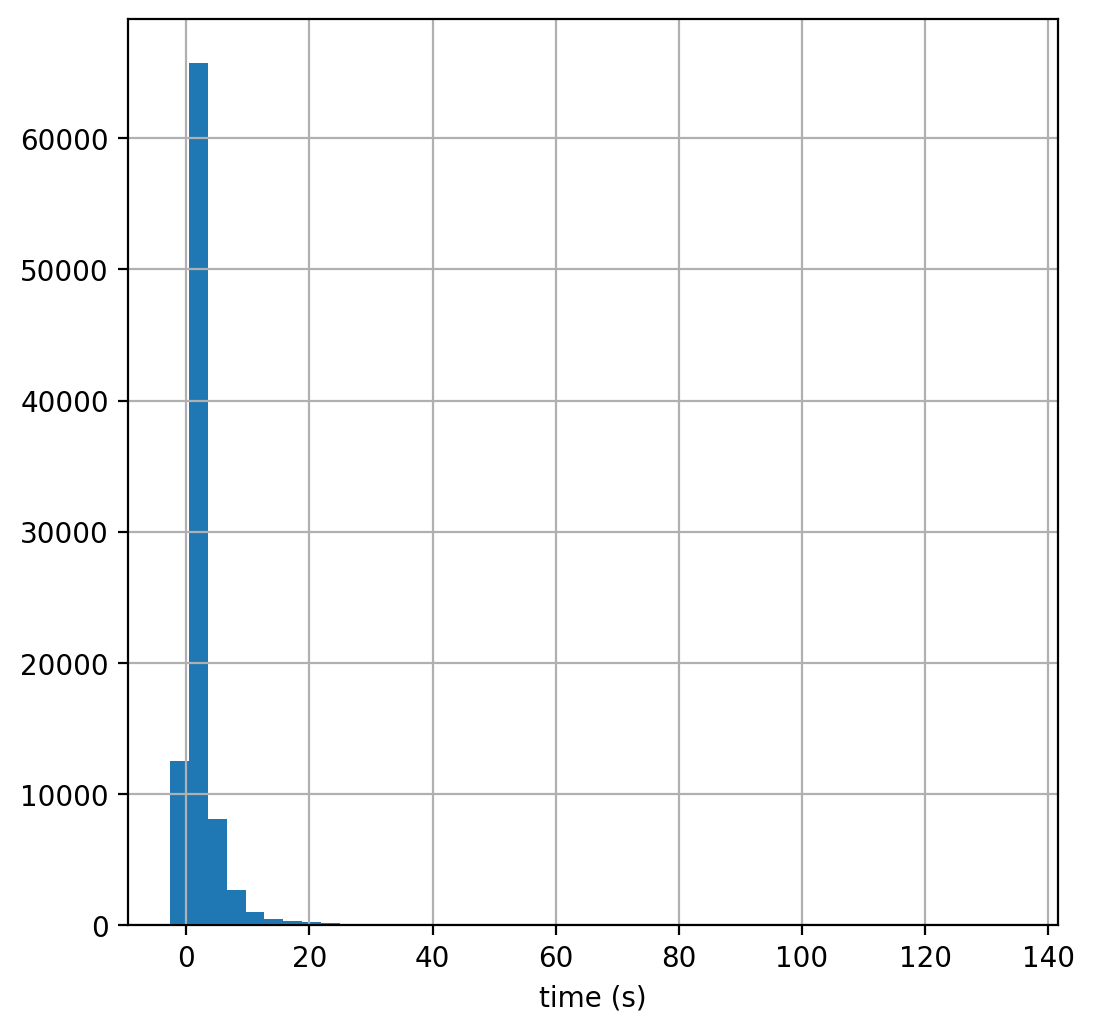} %
\caption{BABEL}
\end{subfigure}

\caption{Motion length distribution}
\label{app:data-motion-length-disstribution}
\end{figure}

\subsection{KIT}
For a fair comparison of the KIT \cite{plappert2016kit} dataset, we use the same pipeline and preprocessing as the TEMOS \cite{petrovich22temos}, using publicly available code.

\section{Appendix B: Derivation}

Variational bound in diffusion model \citep{sohl2015deep, ho2020denoising} is written as:

\begin{equation}
\begin{split}
\mathbb{E} &\left[ -\log p_{\theta} (\boldsymbol{M}_{0}) \right] \leq \mathbb{E}_{q} \left[ -\log \frac{p_{\theta}(\boldsymbol{M}_{0:T})}{q(\boldsymbol{M}_{1:T}|\boldsymbol{M}_{0})} \right] \\
&= \mathbb{E}_{q} \left[ -\log p(\boldsymbol{M}_{T}) - \sum_{t=2}^{T} \log \frac{p_{\theta}(\boldsymbol{M}_{t-1}|\boldsymbol{M}_{t})}{q(\boldsymbol{M}|\boldsymbol{M}_{t-1})} \right] \\ &:= L
\end{split}
\end{equation}

This can be rewritten as:

\begin{equation}
\begin{aligned}
L &=\mathbb{E}_{q}\left[-\log \frac{p_{\theta}\left(\boldsymbol{M}_{0:T}\right)}{q\left(\boldsymbol{M}_{1:T} | \boldsymbol{M}_{0}\right)}\right] \\
&=\mathbb{E}_{q}\left[-\log p\left(\boldsymbol{M}_{T}\right)-\sum_{t=2}^T \log \frac{p_{\theta}\left(\boldsymbol{M}_{t-1} | \boldsymbol{M}_{t}\right)}{q\left(\boldsymbol{M}_{t} | \boldsymbol{M}_{t-1}\right)}\right] \\
&=\mathbb{E}_{q}\Biggr[ -\log p\left(\boldsymbol{M}_{T}\right)-\sum_{t=2}^{T} \log \frac{p_{\theta}\left(\boldsymbol{M}_{t-1} | \boldsymbol{M}_{t}\right)}{q\left(\boldsymbol{M}_{t} | \boldsymbol{M}_{t-1}\right)} \\ &\quad -\log \frac{p_{\theta}\left(\boldsymbol{M}_{0} | \boldsymbol{M}_{1}\right)}{q\left(\boldsymbol{M}_{1} | \boldsymbol{M}_{0}\right)} \Biggr] \\
&=\mathbb{E}_{q}\Biggr[-\log p\left(\boldsymbol{M}_{T}\right) \\ &\quad -\sum_{t=2}^{T} \log \frac{p_{\theta}\left(\boldsymbol{M}_{t-1} | \boldsymbol{M}_{t}\right)}{q\left(\boldsymbol{M}_{t-1} | \boldsymbol{M}_{t}, \boldsymbol{M}_{0}\right)} \cdot \frac{q\left(\boldsymbol{M}_{t-1} | \boldsymbol{M}_{0}\right)}{q\left(\boldsymbol{M}_{t} | \boldsymbol{M}_{0}\right)} \\ &\quad -\log \frac{p_{\theta}\left(\boldsymbol{M}_{0} | \boldsymbol{M}_{1}\right)}{q\left(\boldsymbol{M}_{1} | \boldsymbol{M}_{0}\right)}\Biggr] \\
&=\mathbb{E}_{q}\Biggr[-\log \frac{p\left(\boldsymbol{M}_{T}\right)}{q\left(\boldsymbol{M}_{T} | \boldsymbol{M}_{0}\right)}-\sum_{t=2}^{T} \log \frac{p_{\theta}\left(\boldsymbol{M}_{t-1} | \boldsymbol{M}_{t}\right)}{q\left(\boldsymbol{M}_{t-1} | \boldsymbol{M}_{t}, \boldsymbol{M}_{0}\right)} \\ &\quad -\log p_{\theta}\left(\boldsymbol{M}_{0} | \boldsymbol{M}_{1}\right)\Biggr] \\
&=\mathbb{E}_{q}\Biggr[\underbrace{D_{\mathrm{KL}}\left(q\left(\boldsymbol{M}_{T} \| \boldsymbol{M}_{0}\right) \| p\left(\boldsymbol{M}_{T}\right)\right)}_{L_{T}} \\ &\quad + \sum_{t=2}^T \underbrace{D_{\mathrm{KL}}\left(q\left(\boldsymbol{M}_{t-1} \| \boldsymbol{M}_{t}, \boldsymbol{M}_{0}\right) \| p_{\theta}\left(\boldsymbol{M}_{t-1} \| \boldsymbol{M}_{t}\right)\right)}_{L_{t-1}} \\ &\quad \underbrace{-\log p_{\theta}\left(\boldsymbol{M}_{0} \| \boldsymbol{M}_{1}\right)}_{L_{0}}\Biggr]\\
\end{aligned}
\end{equation}

In DDPM \cite{ho2020denoising} setting, $L_{T}$ is set to constant since the forward process variance $\beta_{t}$ is assumed to be fixed. \citet{ho2020denoising} proposed a parameterization to represent the mean function $\boldsymbol{\mu}_{\theta}(\boldsymbol{M}_{t},t)$ with $p_{\theta}(\boldsymbol{M}_{t-1}|\boldsymbol{M}_{t}) = \mathcal{N}(\boldsymbol{M}_{t-1};\boldsymbol{\mu}_{\theta}(\boldsymbol{M}_{t}, t), \sigma_{t}^2 \boldsymbol{I})$. Using this, $L_{t-1}$ is written as:

\begin{equation}
L_{t-1}=\mathbb{E}_{q}\left[\frac{1}{2 \sigma_{t}^{2}}\left\|\tilde{\boldsymbol{\mu}}_{t}\left(\boldsymbol{M}_{t}, \boldsymbol{M}_{0}\right)-\boldsymbol{\mu}_{\theta}\left(\boldsymbol{M}_{t}, t\right)\right\|^{2}\right]+C
\end{equation} where $C$ is a constant.

Reparameterizing the equation,
\begin{equation}
q\left(\boldsymbol{M}_{t}|\boldsymbol{M}_{0}\right)=\mathcal{N}\left(\boldsymbol{M}_{t} ; \sqrt{\bar{\alpha}_{t}} \boldsymbol{M}_{0},\left(1-\bar{\alpha}_{t}\right) \mathbf{I}\right)
\end{equation} $\boldsymbol{M_{t}}$ can be rewritten as:
\begin{equation}
    \boldsymbol{M}_{t}(\boldsymbol{M}_{0}, \boldsymbol{\epsilon}) = \sqrt{\bar{\alpha}_{t}} \boldsymbol{M}_{0} + \sqrt{1 - \bar{\alpha}_{t}} \boldsymbol{\epsilon}
\end{equation} where $\boldsymbol{\epsilon} \sim \mathcal{N}(\boldsymbol{0}, \boldsymbol{I})$, $\alpha_{t} = 1 - \beta_{t}$, and $\bar{\alpha}_{t} = \prod_{s=1}^{t} \alpha_{s}$.

If the forward process posterior is conditioned on $\boldsymbol{M}_{0}$, it is tractable:

\begin{equation}
q\left(\boldsymbol{M}_{t-1} | \boldsymbol{M}_{t}, \boldsymbol{M}_{0}\right)=\mathcal{N}\left(\boldsymbol{M}_{t-1} ; \tilde{\boldsymbol{\mu}}_{t}\left(\boldsymbol{M}_{t}, \boldsymbol{M}_{0}\right), \tilde{\beta}_{t} \mathbf{I}\right)
\end{equation} where,
$$\quad \tilde{\boldsymbol{\mu}}_{t}\left(\boldsymbol{M}_{t}, \boldsymbol{M}_{0}\right)=\frac{\sqrt{\bar{\alpha}_{t-1}} \beta_{t}}{1-\bar{\alpha}_{t}} \boldsymbol{M}_{0}+\frac{\sqrt{\alpha_{t}}\left(1-\bar{\alpha}_{t-1}\right)}{1-\bar{\alpha}_{t}} \boldsymbol{M}_{t} \quad$$ and $$\quad \tilde{\beta}_{t}=\frac{1-\bar{\alpha}_{t-1}}{1-\bar{\alpha}_{t}} \beta_{t}$$

Combined these, $L_{t-1}$ is written as:
\begin{equation}
\begin{aligned}
&L_{t-1}-C \\
&=\mathbb{E}_{\boldsymbol{M}_{0}, \boldsymbol{\epsilon}}\left[\frac{\beta_{t}^{2}}{2 \sigma_{t}^{2} \alpha_{t}\left(1-\bar{\alpha}_{t}\right)}\left\|\boldsymbol{\epsilon}-\boldsymbol{\epsilon}_{\theta}\left(\sqrt{\bar{\alpha}_{t}} \boldsymbol{M}_{0}+\sqrt{1-\bar{\alpha}_{t}} \boldsymbol{\epsilon}, t\right)\right\|^{2}\right]
\end{aligned}
\end{equation}

In DDPM \cite{ho2020denoising}, they set the last term of the reverse process $L_{0}$ to an independent discrete decoder since the image is distributed in discrete values ($0,1,\ldots,255$). However, we use the same form of $L_{t-1}$ derived above for the $L_{0}$ since we represent motion in continuous values for all time-steps.

\citet{ho2020denoising} empirically found that ignoring weight $\frac{\beta_{t}^{2}}{2 \sigma_{t}^{2} \alpha_{t}\left(1-\bar{\alpha}_{t}\right)}$ improves the sample quality. Finally, the objective function of DDPM is summarized as:
\begin{equation}
L_{\text {simple }}(\theta)=\mathbb{E}_{t, \boldsymbol{M}_{0}, \epsilon}\left[\left\|\boldsymbol{\epsilon}-\boldsymbol{\epsilon}_{\theta}\left(\sqrt{\bar{\alpha}_{t}} \boldsymbol{M}_{0}+\sqrt{1-\bar{\alpha}_{t}} \boldsymbol{\epsilon}, t\right)\right\|^{2}\right]
\end{equation}

Unlike DDPM \cite{ho2020denoising}, which ignored $L_{T}$ in training as the forward process variances $\beta_{t}$ are fixed to constants, Improved DDPM \cite{nichol2021improved} suggests learning the variance as well. Thus, variational lower bound (VLB) in the objective function is summarized as:

\begin{equation}
    L_{\textrm{hybrid}} = L_{\textrm{simple}} + \lambda L_{\textrm{vlb}}
\end{equation} where $L_{\textrm{vlb}} = L_{0} + L_{1} + \cdots + L_{T-1} + L_{T}$

\section{Appendix C: Evaluation Details}

\subsubsection{APE and AVE}
We compute APE and AVE metrics as explained in TEMOS:

\begin{equation}
    \text{APE}[j] = \frac{1}{NF} \sum_{n \in N} \sum_{f \in F} \left\lVert \boldsymbol{M}_{f}[j] - \hat{\boldsymbol{M}}_{f}[j] \right\rVert_2
\end{equation}

\begin{equation}
    \text{AVE}[j] = \frac{1}{N} \sum_{n \in N} \left\lVert \sigma[j] - \hat{\sigma}[j] \right\rVert
\end{equation} where,
\begin{equation} \label{eq:jd}
    \sigma[j] = \frac{1}{F-1} \sum_{f \in F}  \left( \boldsymbol{M}_f[j] - \tilde{\boldsymbol{M}}_f[j] \right)^2 \in \mathbb{R}^3
\end{equation} where $\boldsymbol{M}$ represents motion sequence, $\tilde{\boldsymbol{M}}$ denotes the average of the joint values over the temporal axis, $j$ represents joint index and $f$ indicates frame index. $N$ means the total number of samples. Equation~\ref{eq:jd} is also used to compute joint diversity.

\subsubsection{Multimodality}
Multimodality \cite{guo2020action2motion} measures the motion diversity generated from each text. For the number of text annotations $C$, two sets of $S_{l}$ samples are generated per each annotation. Then, feature vectors are extracted from the samples in each subset using a feature extractor. Multimodality is computed by:

\begin{equation}
    \textrm{Multimodality} = \frac{1}{C \times S_{l}} \sum_{c=1}^{C} \sum_{i=1}^{S_{l}} \lVert \boldsymbol{v}_{c,i} - \boldsymbol{v}_{c,i}^{\prime} \rVert
\end{equation}

We set $S_{l}=10$ in experiments.

\section{Appendix D: Hyperparameters}

\subsection{Hyperparameters of FLAME}

Hyperparameters used in FLAME for HumanML3D and BABEL are listed in Table~\ref{tab:hyperp_hml3d_babel}. FLAME hyperparameters used for the KIT data are listed in Table~\ref{tab:hyperp_kit}. Exponential moving average (EMA) is employed to stabilize training process.

\begin{table}[h]
    \centering
    \begin{tabular}{c|c} \hline
        Hyperparameter & HumanML3D and BABEL  \\ \hline
        Optimizer & AdamW \\
        Learning Rate & 0.0001 \\
        Weight Decay & 0.0001 \\
        $\beta_1$ & 0.9 \\
        $\beta_2$ & 0.999 \\
        \hline
        Diffusion Steps & 1,000 \\
        $\beta$-schedule & cosine \\
        Schedule Sampler & uniform \\
        Motion Dim & 147 \\
        Motion/Language Latent Dim & 768 \\
        Maximum Motion Length & 471 \\
        Maximum Text Length & 20 words \\
        Text Replacement with Null $p$ & 0.25 \\
        Num Heads & 8 \\
        Feedforward Dim & 2,048 \\
        Dropout & 0.1 \\
        Num Layers & 8 \\
        Guidance Scale & 8.0 \\
        EMA Update & 10 \\
        EMA Decay & 0.99 \\
        \hline
    \end{tabular}
    \caption{Hyperparameters for the HumanML3D and BABEL}
    \label{tab:hyperp_hml3d_babel}
\end{table}

\begin{table}[h]
    \centering
    \begin{tabular}{c|c} \hline
        Hyperparameter & KIT  \\ \hline
        Optimizer & AdamW \\
        Learning Rate & 0.0001 \\
        Weight Decay & 0.0001 \\
        $\beta_1$ & 0.9 \\
        $\beta_2$ & 0.999 \\
        \hline
        Diffusion Steps & 1,000 \\
        $\beta$-schedule & cosine \\
        Schedule Sampler & uniform \\
        Motion Dim & 64 \\
        Motion/Language Latent Dim & 512 \\
        Maximum Motion Length & 1,400 \\
        Maximum Text Length & 20 words \\
        Text Replacement with Null $p$ & 0.25 \\
        Num Heads & 8 \\
        Feedforward Dim & 2,048 \\
        Dropout & 0.1 \\
        Num Layers & 8 \\
        Guidance Scale & 8.0 \\
        EMA Update & 10 \\
        EMA Decay & 0.99 \\
        \hline
    \end{tabular}
    \caption{Hyperparameters for the KIT}
    \label{tab:hyperp_kit}
\end{table}

\subsection{Hyperparameters of Motion CLIP}

To extract high-level features from motion and text, we train a separate model in a contrastive manner. This model follows the architecture of CLIP \cite{radford2021learning}. Table~\ref{tab:hyperp_mclip} lists the hyperparameters used to train this model.

\begin{table}[h]
    \centering
    \begin{tabular}{c|c} \hline
        Hyperparameter & Motion CLIP  \\ \hline
        Optimizer & AdamW \\
        Learning Rate & 0.0001 \\
        Weight Decay & 0.0001 \\
        $\beta_1$ & 0.9 \\
        $\beta_2$ & 0.999 \\
        \hline
        Motion Dim & 147 \\
        Motion Encoder Feedforward Dim & 768 \\
        Motion Encoder Latent Dim & 512 \\
        Motion Encoder Num Layers & 6 \\
        Motion Encoder Num Heads & 8 \\
        Motion Encoder Dropout & 0.1 \\
        Maximum Motion Length & 471 \\
        \hline
        Text Encoder Latent Dim & 512 \\
        \hline
        Loss & InfoNCE \\
        $\tau$ (Temperature) & 0.07 \\
        \hline
    \end{tabular}
    \caption{Hyperparameters for Motion CLIP}
    \label{tab:hyperp_mclip}
\end{table}

\section{Appendix E: Additional Experiments}

\subsection{Comparison of Sampling Methods}

\begin{table}[h]
\centering
\begin{subtable}[h]{0.9\columnwidth}
\centering
\resizebox{\columnwidth}{!}{%
\begin{tabular}{c|cccccc}
\hline
               &        &         &        & \multicolumn{3}{c}{R-Precision} \\ \cline{5-7} 
Sampling Steps & mCLIP & FD     & MID    & Top-1     & Top-2    & Top-3    \\ \hline
1,000          & 0.297  & 21.152  & 29.935 & 0.513    & 0.673   & 0.749   \\
500            & 0.296  & 21.481  & 29.767 & 0.508    & 0.672   & 0.740   \\
250            & 0.296  & 22.459  & 29.101 & 0.507    & 0.666   & 0.747   \\
100            & 0.295  & 22.980  & 29.113 & 0.501    & 0.668   & 0.740   \\
50             & 0.292  & 23.014  & 29.098 & 0.506    & 0.662   & 0.734   \\
25             & 0.293  & 23.447  & 29.035 & 0.509    & 0.666   & 0.746   \\
10             & 0.288  & 29.829  & 28.802 & 0.481    & 0.635   & 0.710   \\
7              & 0.271  & 54.495  & 28.287 & 0.405    & 0.555   & 0.633   \\
5              & 0.223  & 107.864 & 26.057 & 0.306    & 0.439   & 0.519   \\ \hline
\end{tabular}%
}
\caption{DDPM Sampling}
\end{subtable}

\begin{subtable}[h]{0.9\columnwidth}
\centering
\resizebox{\columnwidth}{!}{
\begin{tabular}{c|cccccc}
\hline
                &        &         &        & \multicolumn{3}{c}{R-Precision} \\ \cline{5-7} 
Sampling Steps & mCLIP & FD     & MID    & Top-1     & Top-2    & Top-3    \\ \hline
1,000          & 0.298 & 22.063  & 29.905 & 0.513 & 0.666 & 0.744 \\
500            & 0.298 & 23.025  & 29.856 & 0.516 & 0.669 & 0.747 \\
250            & 0.298 & 23.250  & 29.831 & 0.508 & 0.672 & 0.741 \\
100            & 0.297 & 23.255  & 29.658 & 0.506 & 0.669 & 0.749 \\
50             & 0.295 & 23.624  & 29.662 & 0.519 & 0.672 & 0.748 \\
25             & 0.293 & 25.721  & 29.701 & 0.510 & 0.666 & 0.741 \\
10             & 0.285 & 46.521  & 28.758 & 0.460 & 0.617 & 0.700 \\
7              & 0.260 & 103.748 & 27.730 & 0.367 & 0.519 & 0.602 \\
5              & 0.232 & 178.007 & 25.748 & 0.279 & 0.401 & 0.480 \\ \hline
\end{tabular}%
}
\caption{DDIM Sampling}
\end{subtable}
\caption{Quantitative results on different sampling steps by DDPM and DDIM sampling.}
\label{app:sampling-ddpm-ddim}
\end{table}

With FLAME, we can sample from different strategies: DDPM \cite{ho2020denoising} and DDIM \cite{song2020denoising}. Table~\ref{app:sampling-ddpm-ddim} presents the benchmark for DDPM and DDIM sampling.

\subsection{Benchmark on Different Guidance Scale}
We use classifier-free guidance \cite{ho2021classifier} to perform text-conditional generation in FLAME. To show the effects of varying guidance scales (strength), we conduct sampling on various guidance scales. Table~\ref{tab:diffusion-guidance_scale} presents the result.

\begin{table}[h]
\centering
\resizebox{\columnwidth}{!}{%
\begin{tabular}{c|cccccc}
\hline
               &        &         &        & \multicolumn{3}{c}{R-Precision} \\ \cline{5-7} 
Guidance Scale & mCLIP & FD     & MID    & Top-1     & Top-2    & Top-3    \\ \hline
4           & 0.296 & 20.771  & 29.899 & 50.988 & 66.909 & 74.439 \\
8           & 0.297 & 20.752  & 30.014 & 51.254 & 67.298 & 74.935 \\
12          & 0.297 & 20.961  & 30.006 & 51.346 & 67.370 & 75.549 \\
16          & 0.298 & 21.088  & 29.684 & 51.461 & 66.569 & 74.812 \\
32          & 0.296 & 21.046  & 29.370 & 50.132 & 67.188 & 74.520 \\
64          & 0.295 & 22.034  & 29.592 & 50.505 & 65.967 & 74.115 \\
\hline
\end{tabular}%
}
\caption{Quantitative results on different guidance scale.}
\label{tab:diffusion-guidance_scale}
\end{table}
 
\subsection{Additional Qualitative Results}

\begin{figure}[h]
\captionsetup[subfigure]{labelformat=empty}
\centering

\begin{subfigure}[b]{0.9\columnwidth}
\centering
\includegraphics[width=\columnwidth]{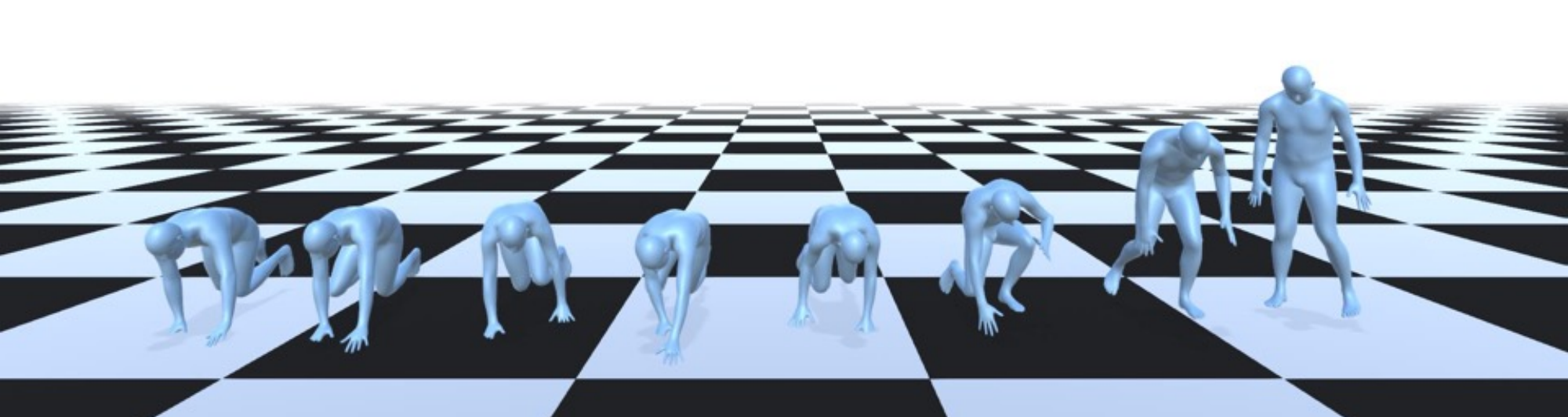} %
\caption{\textit{``A person crawls, and then stands up."}}
\end{subfigure}

\begin{subfigure}[b]{0.9\columnwidth}
\centering
\includegraphics[width=\columnwidth]{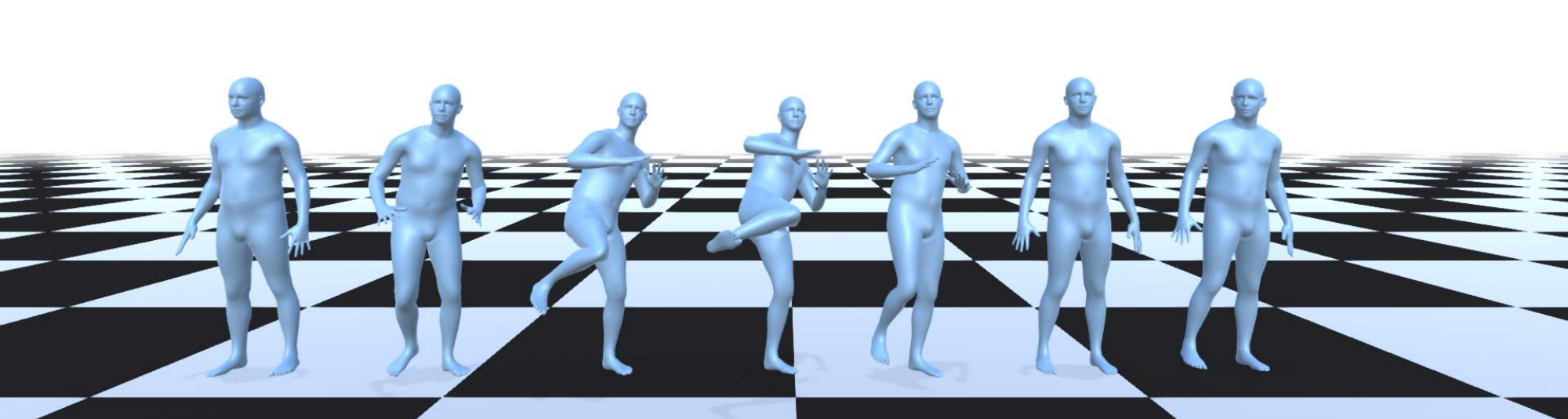} %
\caption{\textit{``A person kicks with his right leg."}}
\end{subfigure}

\begin{subfigure}[b]{0.9\columnwidth}
\centering
\includegraphics[width=\columnwidth]{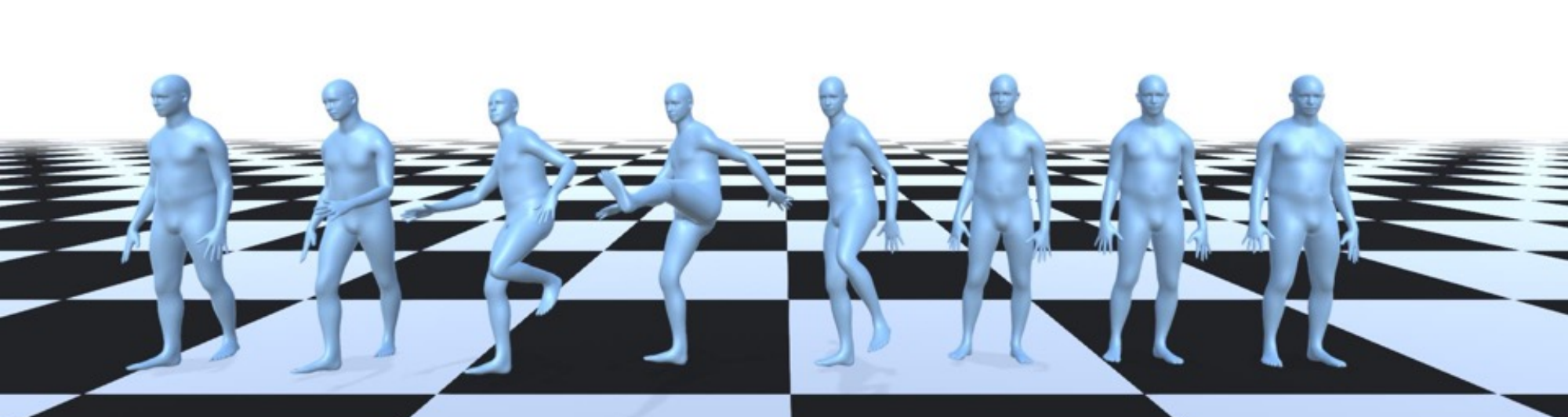} %
\caption{\textit{``A person kicks with his left leg."}}
\end{subfigure}

\begin{subfigure}[b]{0.9\columnwidth}
\centering
\includegraphics[width=\columnwidth]{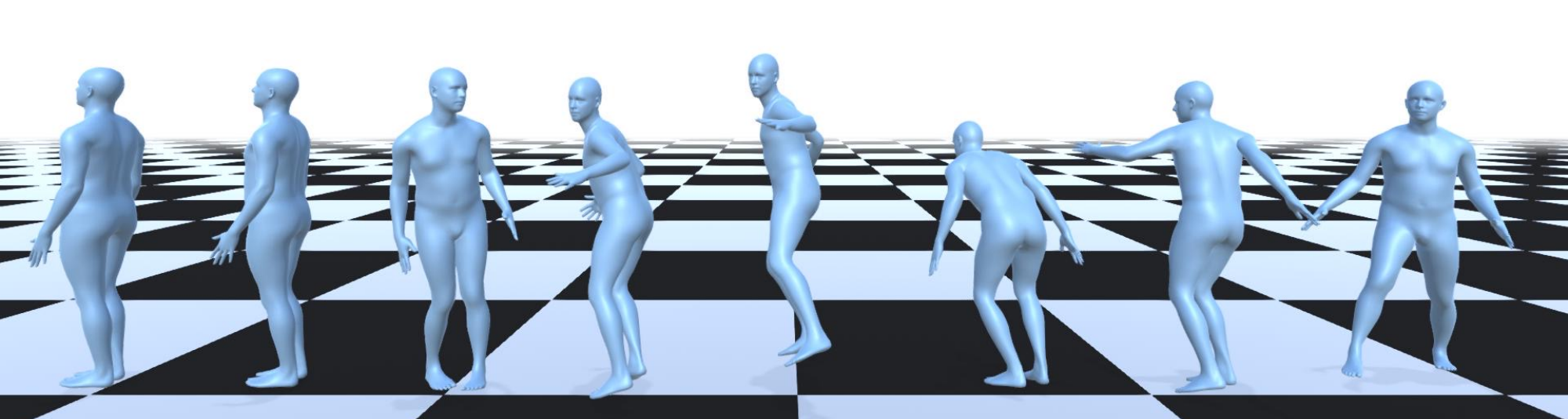} %
\caption{\textit{``A person jumps and spins in the air 360 degree counter-clockwise."}}
\end{subfigure}

\begin{subfigure}[b]{0.9\columnwidth}
\centering
\includegraphics[width=\columnwidth]{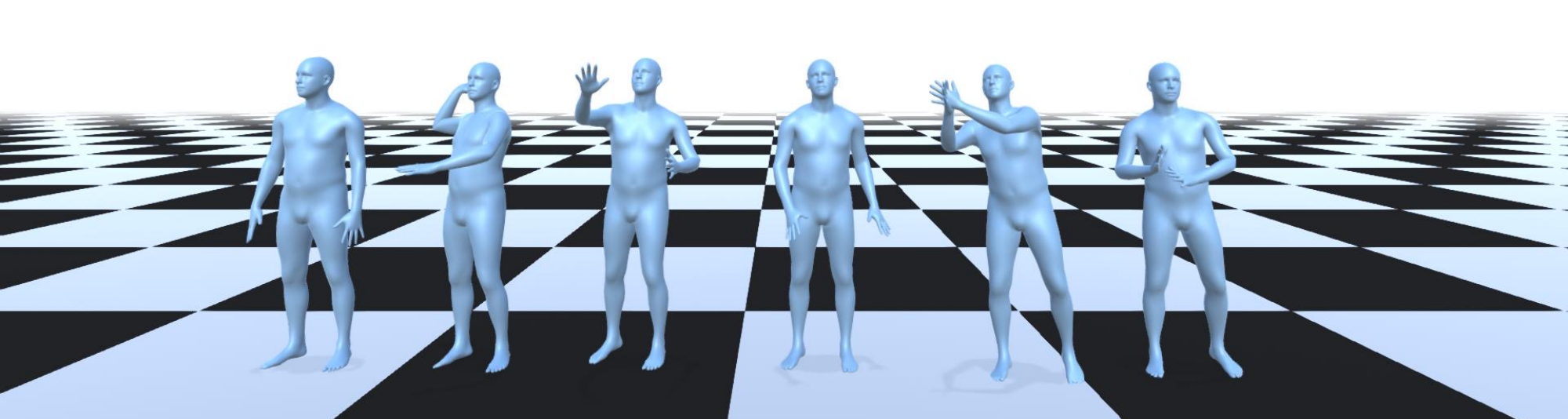} %
\caption{\textit{``A person throws an object with right hand and catches an object with both hands."}}
\end{subfigure}

\caption{Text-to-motion synthesis examples.}
\label{app:text-to-motion}
\end{figure}

\begin{figure}[h]
\captionsetup[subfigure]{labelformat=empty}
\centering

\begin{subfigure}[b]{0.9\columnwidth}
\centering
\includegraphics[width=\columnwidth]{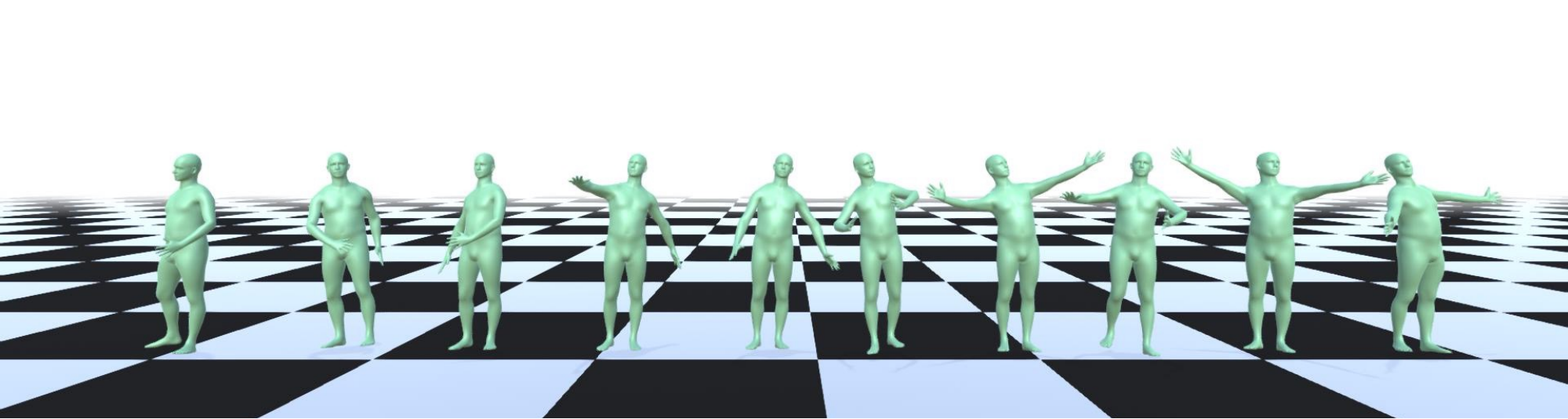} %
\caption{Reference motion}
\end{subfigure}

\begin{subfigure}[b]{0.9\columnwidth}
\centering
\includegraphics[width=\columnwidth]{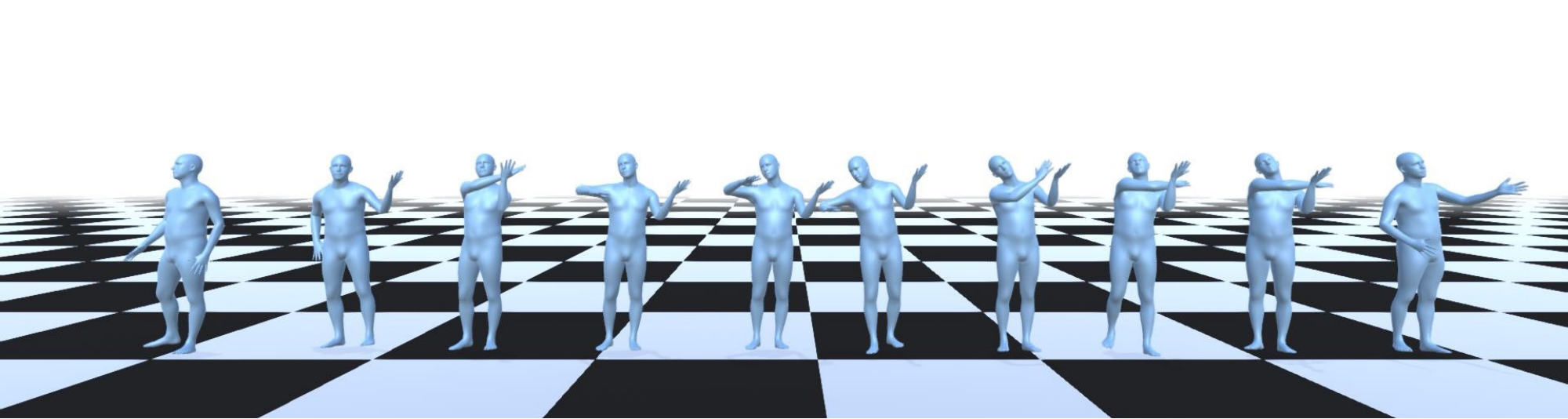} %
\caption{Editing prompt: \textit{``A person is playing a violin.''}}
\end{subfigure}

\begin{subfigure}[b]{0.9\columnwidth}
\centering
\includegraphics[width=\columnwidth]{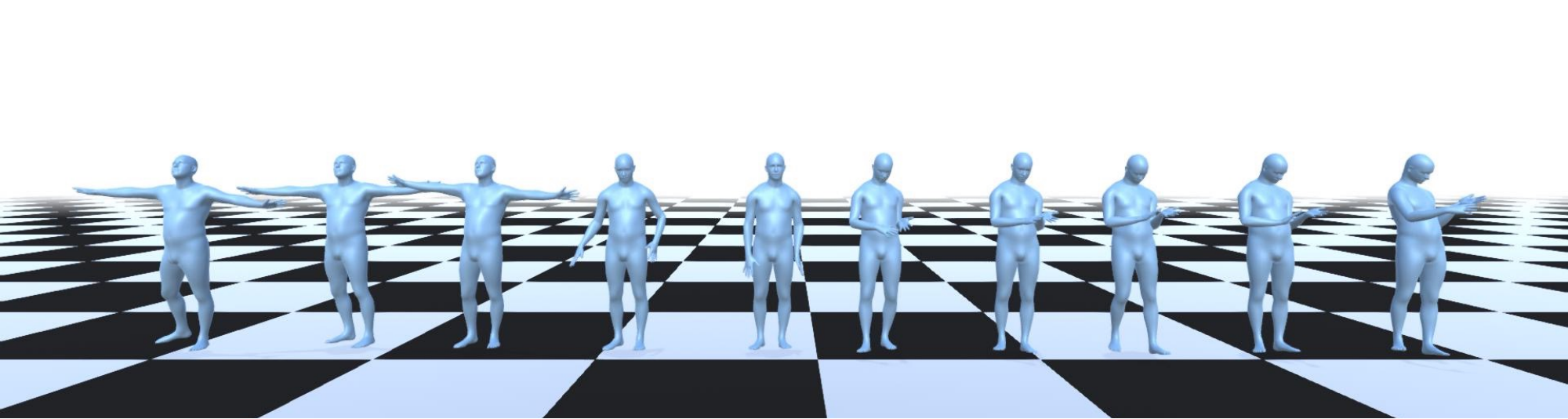} %
\caption{Editing prompt: \textit{``A person makes a phone call.''}}
\end{subfigure}

\begin{subfigure}[b]{0.9\columnwidth}
\centering
\includegraphics[width=\columnwidth]{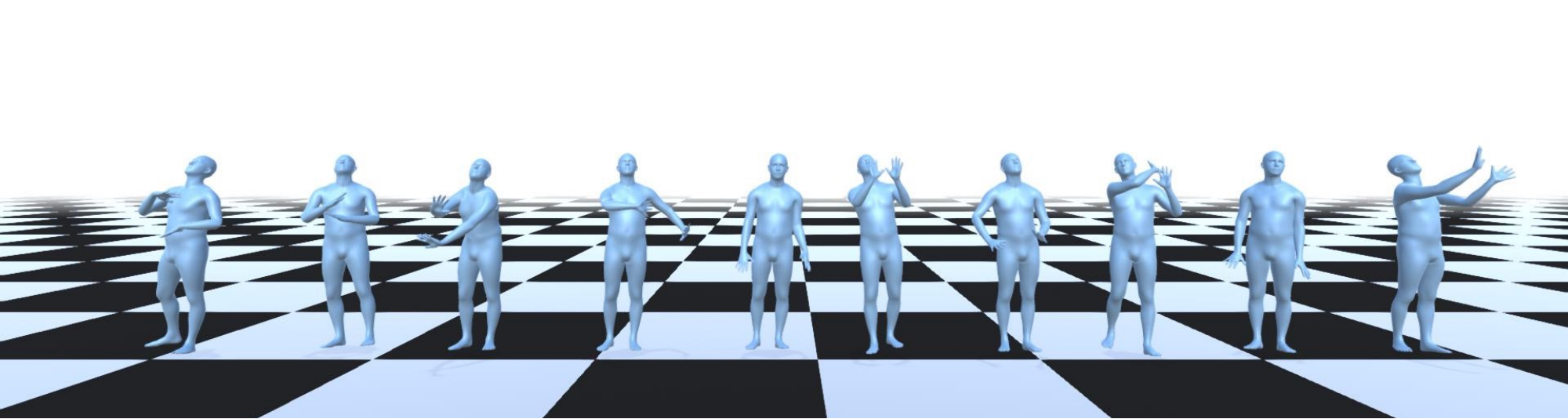} %
\caption{Editing prompt: \textit{``A person throws and catches a ball''}}
\end{subfigure}

\caption{Free-form language-based motion editing example. Model is allowed to edit upper body parts including the root (pelvis) joint.}
\label{app:motion-editing}
\end{figure}

Additional text-to-motion synthesis and editing examples are provided in Figure~\ref{app:text-to-motion} and Figure~\ref{app:motion-editing} respectively.

\nobibliography{aaai23}